\pdfoutput=1
\documentclass[journal]{IEEEtran}%
\IEEEoverridecommandlockouts
\usepackage{cite}
\usepackage{amsmath,amssymb,amsfonts}
\usepackage{amsthm,bm}
\usepackage{mathrsfs}
\usepackage{colortbl}
\usepackage{hhline}
\usepackage{graphicx}
\usepackage{subcaption}
\usepackage{textcomp}
\usepackage{stfloats}
\usepackage{pifont}
\usepackage{enumerate}
\usepackage{booktabs}
\usepackage{url}
\usepackage{mathtools}
\usepackage{cuted}
\usepackage{array}  %表格上下左右居中

\usepackage {threeparttable}
\usepackage{tablefootnote}
\usepackage{multirow}
\usepackage{colortbl}
\def\BibTeX{{\rm B\kern-.05em{\sc i\kern-.025em b}\kern-.08em
    T\kern-.1667em\lower.7ex\hbox{E}\kern-.125emX}}

\usepackage{stackengine}
\usepackage{enumerate}
\usepackage{booktabs}
\usepackage{subfig}
\usepackage{multicol}
\usepackage[noend]{algpseudocode}
\usepackage{pgfplots}
\usepackage{pgfplotstable}
\usepackage[ruled,linesnumbered]{algorithm2e} 
\usepackage{amssymb,amsmath,caption}
\usepackage[hang,flushmargin]{footmisc}
\usepackage{lipsum}
\makeatletter
\newcommand{\algorithmfootnote}[2][\footnotesize]{%
  \let\old@algocf@finish\@algocf@finish% Store algorithm finish macro
  \def\@algocf@finish{\old@algocf@finish% Update finish macro to insert "footnote"
    \leavevmode\rlap{\begin{minipage}{\linewidth}
    #1#2
    \end{minipage}}%
  }%
}
\SetKwRepeat{Do}{do}{while}
\usepackage{setspace}
\usepackage{color}
\usepackage{tabularray}
\usepackage{pifont} 
\makeatletter
\def\BState{\State\hskip-\ALG@thistlm}
\makeatother

\algnewcommand\algorithmicforeach{\textbf{for each}}
\algdef{S}[FOR]{ForEach}[1]{\algorithmicforeach\ #1\ \algorithmicdo}
\begin{document}
% \DefineFNsymbols{circled}{{\ding{192}}{\ding{193}}{\ding{194}}
% {\ding{195}}{\ding{196}}{\ding{197}}{\ding{198}}{\ding{199}}
% {\ding{200}}{\ding{201}}}
\title{ Channel-adaptive Cross-modal Generative Semantic Communication  for  Point Cloud   Transmission
}
\author{Wanting Yang, Zehui Xiong, \textit{Senior Member, IEEE}, Qianqian Yang,  \textit{Member, IEEE}, Ping Zhang, \textit{Fellow, IEEE}, Mérouane Debbah, \textit{Fellow, IEEE}, Rahim Tafazolli, \textit{Fellow, IEEE}
\thanks{
% Wanting Yang is with the Pillar
% of Engineering Systems and Design, Singapore University of Technology and Design, Singapore (e-mail: wanting\_yang@sutd.edu.sg);
% Zehui Xiong is with the Pillar
% of Information Systems Technology and Design, Singapore University of
% Technology and Design, Singapore (e-mail: zehui\_xiong@sutd.edu.sg); 
% Qianqian Yang is with College of Information Science and Electronic Engineering, Zhejiang University, Hangzhou, China (email: qianqianyang20@zju.edu.cn);
% Ping Zhang is with the School of Information and Communication Engineering, Beijing University of Posts and Telecommunications, and also with
% the State Key Laboratory of Networking and Switching Technology, Beijing
% 100876, China (email: pzhang@bupt.edu.cn);
% Merouane Debbah is with  KU 6G Research Center, Department of Computer and Information Engineering, Khalifa University, Abu Dhabi 127788, UAE (email: merouane.debbah@ku.ac.ae) and also with CentraleSupelec, University Paris-Saclay, 91192 Gif-sur-Yvette, France;
% Rahim Tafazolli is with the Institute
% for Communication Systems (ICS), 5GIC \& 6GIC, University of Surrey,
% Guildford, Surrey, GU2 7XH, U.K. (email:
% r.tafazolli@surrey.ac.uk).
}
}

\makeatletter
\setlength{\@fptop}{0pt}
\makeatother

\maketitle

\vspace{-1.8cm}
\begin{abstract}
With the rapid development of autonomous driving and extended reality, efficient transmission of point clouds (PCs) has become increasingly important.  In this context, we propose a novel channel-adaptive cross-modal generative semantic communication (SemCom) for PC transmission, called GenSeC-PC. GenSeC-PC employs a semantic encoder that fuses images and point clouds, where images serve as \textit{non-transmitted} side information. Meanwhile, the decoder is built upon the backbone of PointDif. Such a cross-modal design not only ensures high compression efficiency but also delivers superior reconstruction performance compared to PointDif. Moreover, to ensure robust transmission and reduce system complexity, we design a streamlined and asymmetric  channel-adaptive joint semantic-channel coding architecture, where only the encoder needs the feedback of average signal-to-noise ratio (SNR) and available bandwidth. In addition, rectified denoising diffusion implicit models is employed to accelerate the decoding process to the millisecond level, enabling real-time PC communication. Unlike existing methods, GenSeC-PC leverages generative priors to ensure reliable reconstruction even from noisy or incomplete source PCs. More importantly, it supports \textit{fully analog} transmission, improving compression efficiency by eliminating the need for error-free side information transmission common in prior SemCom approaches. Simulation results confirm the effectiveness of cross-modal semantic extraction and dual-metric guided fine-tuning, highlighting the framework’s robustness across diverse conditions—including low SNR, bandwidth limitations, varying numbers of 2D images, and previously unseen objects.
\end{abstract}


\newtheorem{definition}{Definition}
\newtheorem{lemma}{Proposition}
\newtheorem{theorem}{Theorem}

\newtheorem{property}{Property}

\vspace{0cm}
\section{Introduction}
Point clouds (PCs) provide a fundamental yet highly efficient representation of 3D objects by capturing their shapes as collections of measured point coordinates. Thanks to their simplicity and strong representational power~\cite{han2017review}, PC processing has made notable progress, with applications in robotics, autonomous driving, and more~\cite{fan2021pstnet}. Meanwhile, the rise of emerging technologies such as the metaverse and digital twins has sparked growing interest in PC transmission, which is essential for high-quality 3D reconstruction in remote scenarios. As a result, the development of efficient transmission solutions has recently gained significant attention as a new direction in both academia and industry.

A series of studies have been conducted on PC compression~\cite{cao20193d}. Nonetheless, most existing methods, such as Octree-based coding~\cite{schnabel2006octree}, primarily rely on the syntactic structure of raw data, which limits compression efficiency and leaves the compressed data size still relatively large.  Meanwhile, the ``cliff effect” prevalent in traditional communication poses additional challenges for  robust and reliable PC transmission~\cite{10679082}. 
To address these issues, semantic communication (SemCom) has been widely recognized as a promising solution.
However, unlike 2D images arranged in a regular grid, 3D PCs are unstructured, unordered, and discretely distributed~\cite{zhou2021adaptive}. This irregularity complicates semantic analysis and efficient semantic extraction. As a result, SemCom for PCs is still in its early stages, lagging behind its counterparts for data from modalities such as vision, audio, and text. 
Current state-of-the-art research in SemCom for PCs generally follows two main directions.

One line of existing work leverages deep learning backbones  for semantic  extraction, serving as the encoder and decoder in JSCC frameworks. Specifically, Han \textit{ et al.} proposed a SemCom paradigm for PC classification based on the Point-BERT backbone~\cite{yu2022point} in \cite{10437861}. Then, for the PC reconstruction, Xie \textit{et al.} and Zhang \textit{et al.} introduced two innovative JSCC-based transmission paradigms. Xie \textit{et al.} utilized PointNet++~\cite{10.5555/3295222.3295263} as the core module for semantic extraction in~\cite{10901573}, while Zhang \textit{et al.} employed Inception-Residual Networks~\cite{10.5555/3298023.3298188} for the same purpose in~\cite{10889738}. Nonetheless, both studies rely on the \textit{shared} side information, such as global semantic representations and the center coordinates of PC subgroups, that must be transmitted in a lossless digital manner to assist in PC  reconstruction at the receiver side.  The associated overhead from digital quantization and channel coding often results in a compression rate that falls short of expectations, especially when compared to JSCC frameworks developed for  structured data modalities.
Moreover, Ibuki \textit{et al.} proposed a rateless deep JSCC for PC, where the graph autoencoder architecture~\cite{9500925} is based on graph neural
networks in~\cite{10906588}. However, the compressed semantic information still requires tens of thousands of symbols to represent a single object in ShapeNet~\cite{chang2015shapenet} dataset. 
Apart from the above, there has also been research focused on PC-based volumetric media~\cite{10622850, huang2023iscom}. 
However, their focus was limited to a single category, specifically human gestures. In this sense, such methods may not generalize well to scenarios involving multiple categories. 
Meanwhile, the above JSCC frameworks face an  inherently limitation stemming from their reliance on the conventional \textit{compression–decompression pipeline}. This paradigm requires the compressed representation to retain all semantic content necessary for accurate reconstruction and implicitly assumes that the input data is complete and noise-free. In practice, however, PCs are often incomplete due to occlusions and environmental interference, which makes these assumptions difficult to satisfy and degrades reconstruction performance.

The other line of research follows the generative SemCom paradigm, where only 2D images are transmitted from the sender, and PCs are reconstructed at the receiver using RGB-depth alignment techniques~\cite{zhang2024semantic, jiang2024large}. Specifically, Zhang~\textit{et al.} utilized both RGB and depth frames as source data in \cite{zhang2024semantic}, and Jiang~\textit{et al.} employed multi-perspective images of selected 3D objects as the source data in \cite{jiang2024large}.
While these approaches can offer higher transmission efficiency, they often fail to preserve fine-grained object details across diverse viewpoints. This limitation primarily arises from the omission of rich 3D semantic content originally embedded in the PC. Meanwhile, factors such as lighting conditions and alignment errors can substantially affect the fidelity of the reconstructed PCs. Furthermore, from a methodological standpoint, these approaches—by projecting PCs into 2D space at the transmitter and reconstructing them at the receiver—essentially bypass the inherent challenges associated with direct PC transmission, thereby reducing the task to a conventional image transmission problem.
In addition, a  multi-modal SemCom approach~\cite{10888756} has been proposed based on a generative model, ${\rm PC}^2$~\cite{melas2023pc2}. This method transmits a 2D image from a specific viewpoint along with a 2D projection of the raw point cloud, rasterized using an orthographic camera. As the semantic extraction focuses on a single view rather than the full 3D structure, the quality of the reconstructed point cloud can only be reliably ensured from that specific perspective. For viewpoints not captured during semantic extraction, the reconstructed point cloud can only be guaranteed to form a 3D structure, while the accuracy of its shape and contour from those perspectives cannot be reliably ensured~\cite{melas2023pc2}.

In response to this gap,  we propose a novel cross-modal generative SemCom framework,  called GenSeC-PC, that  seamlessly integrates generative SemCom with JSCC. At the transmitter side, the semantic encoder is redesigned as a prompt engineering module, which extracts essential semantic information from only a partial subset of the source PC. This enables substantial compression while preserving key structural information. Additionally, multi-view 2D images of the object are introduced as auxiliary input. These images are not transmitted but are used locally to enhance 3D keypoint detection, thereby improving the accuracy and robustness of semantic feature extraction.
At the receiver side, the semantic decoder performs controlled generation using a diffusion-based generative model. Leveraging prior knowledge, it reconstructs the complete PC from the extracted semantics, allowing the model to infer global structure from partial observations. This effectively reduces the impact of incomplete or noisy input data on reconstruction quality. Furthermore, the inference process is significantly streamlined to lower decoding complexity, enabling real-time communication without sacrificing reconstruction fidelity. The key contributions of our work have been highlighted as below.
\begin{itemize}
    \item We design a cross-modal semantic encoder that integrates structural features from 2D images with both local and global features from 3D PCs, enabling more comprehensive semantic representation. The resulting semantic representation can be transmitted entirely in a lossy analog manner. This differs from prior SemCom frameworks based solely on JSCC, which require digital transmission to achieve lossless transmission of critical information, such as global semantics and key coordinates. Furthermore, the use of a random masking mechanism during semantic extraction enhances the model’s robustness, allowing reliable reconstruction even when the source point cloud is noisy or incomplete.
    \item We design a streamlined and \textit{asymmetric} channel-adaptive JSCC scheme  based on pruning techniques. Only the semantic encoder requires feedback of the average SNR and available bandwidth. It first employs multiple SNR Adaptation Components (SACs) to refine semantic features with SNR-aware modulation. Then, a two-stage rate-oriented compression is performed: multiple Rate Adaptation Components (RACs) adapt features to different rates, followed by a Conditional Multi-Branch Component (CMBC) that removes the need for explicit rate signaling, reducing coordination overhead. This design balances redundancy and reconstruction accuracy.
On the receiver side, the decoder first performs a rate-adaptive semantic decompression that mirrors the encoder’s process, guided by the received feature length. To mitigate semantic degradation caused by feature-dependent variability, multiple Feature Adaptation Components (FACs) are used to enhance the decoded features accordingly.

    \item We conduct the training and simulations using the KeypointNet and ShapeNet dataset. Simulation results show that GenSeC-PC is highly robust to channel variations, different numbers of input images, and unseen objects, while achieving significantly higher spectral efficiency than traditional methods. In addition, an ablation study is conducted to validate the necessity and superiority  of the proposed cross-modal semantic extraction module and dual-metric guided fine-tuning. To mitigate the decoding latency, Denoising Diffusion Implicit Models (DDIM) and Rectified Diffusion (RD) are employed during the inference process. These techniques significantly reduce the decoding time to the millisecond level, enabling practical deployment in real-time communication scenarios while maintaining a favorable trade-off between latency and reconstruction accuracy. 

\end{itemize}

The remainder of this paper is organized as follows. Section~\ref{sec:systemovervive} introduces the overview of  GenSeC-PC framework. The detailed design of each component is presented in Section~\ref{sec:detailed}. Section~\ref{sec:simulation} validates the effectiveness of the proposed approach. Finally, Section~\ref{sec:conclusion} concludes the paper.

\section{System Overview and Metrics}
\label{sec:systemovervive}
\begin{figure*}
    \centering
    \includegraphics[width=1\linewidth]{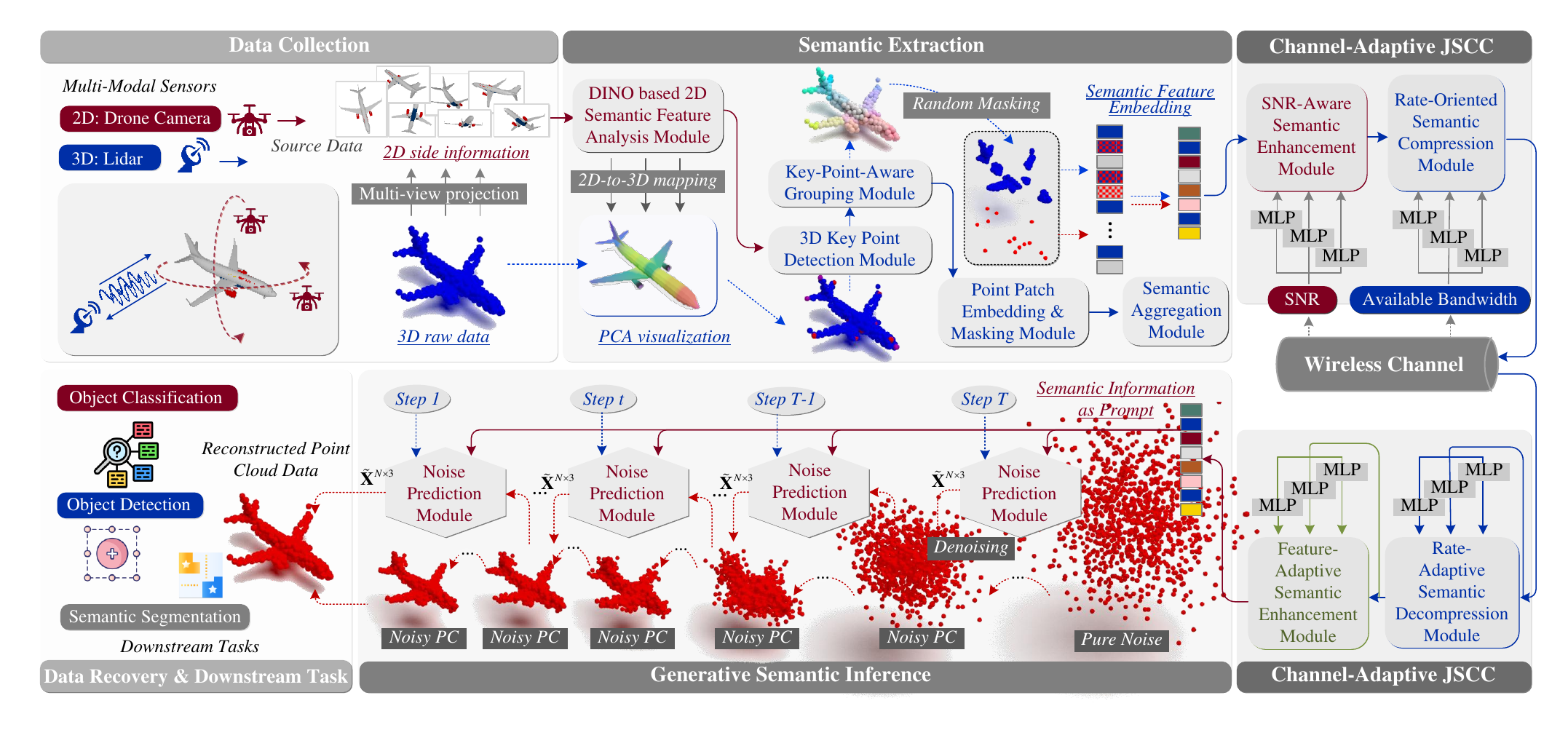}
    \caption{Illustration of channel-adaptive cross-modal GenSeC-PC framework.}
    \label{fig:systemmodel}
    \vspace{-0.3cm}
\end{figure*}

% In this work, we focus on a PC reconstruction task and propose a channel-adaptive cross-modal GenSeC-PC framework, as shown in Fig.~\ref{fig:systemmodel}. In this section, we detail the system overview and the involved communication metrics.

\subsection{System Overview}
In this work, we focus on a PC reconstruction task and propose a channel-adaptive cross-modal GenSeC-PC framework, as shown in Fig.~\ref{fig:systemmodel}.
At the source side, we consider the presence of two types of sensors that cooperatively perform data collection. The first is a 3D sensor, such as LiDAR, used to capture the PC of the target object. The second is a mobile camera, such as one mounted on a drone, which captures multi-view 2D images of the object from different perspectives. The multiple 2D images then serve as side information to assist the holistic semantic extraction from the PC. At the receiver side, we employ a diffusion model as the core of the decoder. Its learned generative capability enables the system to tolerate semantic degradation, arising from acquisition noise or transmission distortions,  while still accurately reconstructing the source PC.

Specifically, leveraging multi-view 2D images, keypoints within the PC are first identified using the B2-3D method~\cite{wimmer2023back}. 
% This method employs a large pre-trained vision encoder, DINO~\cite{oquab2024dinov}, to analyze object shapes in 2D views and back-project the extracted features onto the 3D surface. Keypoint detection is then performed by matching the principal components of the 3D features with a small set of pre-learned keypoint representations.
This design is primarily motivated by the following considerations: (1) Intuitively, keypoints in a PC capture the distinctive contours and shape characteristics of individual objects; (2) Existing keypoint detection methods applied directly to 3D data often suffer from limited training datasets, leading to reduced performance on complex models.
After keypoint detection, the PC is divided into patches using a keypoint-aware grouping module, in line with standard point cloud analysis procedures. To help the backbone network learn richer geometric priors during pre-training, random masking is applied~\cite{zheng2024point}. This enables robust reconstruction at the receiver side, even when the original PC is noisy or incomplete. Finally, semantics extracted from both keypoints and the full PC are combined.
Subsequently, semantic extraction is performed through two sequential modules: the Point Patch Embedding \& Masking Module and the Semantic Aggregation Module. 

The final semantic feature is subsequently fed into a pair of asymmetric channel-adaptive JSCC encoder and decoder. The JSCC encoder incorporates both the average SNR and real-time available bandwidth as the feedback to adaptively refine and extract the most relevant semantic features. 
 The compressed semantic information is then transmitted over the wireless channel.
At the receiver, the received semantic features are first fed into a JSCC decoder, which operates in two stages. First, it performs adaptive semantic decompression based on the length of the received information. Second, it conducts adaptive semantic enhancement guided by the received features themselves, aiming to compensate for the heterogeneous semantic loss across different semantic components caused by transmission. Notably, the JSCC decoder operates without requiring any feedback from the wireless channel.
After that, the recovered semantic feature serves as the prompt for the diffusion-based semantic decoder to guide PC reconstruction. Furthermore, various downstream tasks can be performed based on the reconstructed PC, such as object classification, object detection, and semantic segmentation.

\vspace{-0.3cm}

\subsection{Communication Metrics}
For the considered PC reconstruction task, the communication performance is measured by the similarity between the source point cloud $\mathbf{x}$ and the reconstructed PC $\hat{\mathbf{x}}$ at the receiver. Unlike 2D images with grid-aligned pixels, PCs are unordered, making conventional Mean Squared Error (MSE) an inadequate metric. Let $x \in \mathbf{x}$ denote an arbitrary point in the source PC. More metrics are introduced as follows.
\subsubsection{Mean Squared Error}
MSE computes the average squared distance between corresponding points in two PCs.  The primary role of MSE loss in this work is to guide the training of the noise prediction module in the diffusion-based decoder. It can be expressed as
\begin{equation}
    {\text{MSE}}\left( {{\bf{x}},{\bf{\hat x}}} \right) = \frac{1}{N}\sum\limits_{x \in {\bf{x}}} {\left\| {x - \hat x} \right\|_2^2} ,
\end{equation}
where $N$ represents the total number of the points in a PC.
\subsubsection{Chamfer Distance}
Chamfer Distance (CD) computes the average nearest-neighbor distance between the two PCs in both directions, and it is robust to unordered and varying point densities. It is the most widely used metric in 3D vision tasks and can expressed by 
\begin{equation}
\text{CD}(\mathbf{x}, \hat{\mathbf{x}}) = \frac{1}{|\mathbf{x}|} \sum_{x \in \mathbf{x}} \min_{\hat{x} \in \hat{\mathbf{x}}} \| x - \hat{x} \|_2^2 + \frac{1}{|\hat{\mathbf{x}}|} \sum_{\hat{x} \in \hat{\mathbf{x}}} \min_{x \in \mathbf{x}} \| \hat{x} - x \|_2^2. \label{eq:CD}
\end{equation}
\subsubsection{Hausdorff Distance}
Hausdorff Distance (HD) reflects the worst-case point deviation by computing the maximum nearest-neighbor distance from each point set to the other. Compared to CD, which focuses more on assessing the overall reconstruction quality of PCs, HD is more sensitive to local geometric discrepancies, which can be expressed by
\begin{equation}
\text{HD}(\mathbf{x}, \hat{\mathbf{x}}) = \max \left\{ \sup_{x \in \mathbf{x}} \inf_{\hat{x} \in \hat{\mathbf{x}}} \| x - \hat{x} \|_2,\ \sup_{\hat{x} \in \hat{\mathbf{x}}} \inf_{x \in \mathbf{x}} \| \hat{x} - x \|_2 \right\}.
\end{equation}
\subsubsection{Earth Mover’s Distance}
Earth Mover’s Distance (EMD) provides a more rigorous assessment of PC similarity by computing the minimum cost of transforming one point set into another through a bijective mapping. Compared to CD, which measures average nearest-neighbor distances, EMD captures the global structural alignment more accurately by enforcing one-to-one correspondences. Thus, EMD provides a more precise reflection of slight global displacements, local point aggregations, or partial missing regions within a PC.
% Unlike HD, which focuses on the worst-case deviation, EMD reflects the overall matching cost across the entire PC. 
It can be expressed as 
\begin{equation}
\text{EMD}(\mathbf{x}, \hat{\mathbf{x}}) = \min_{\phi: \mathbf{x} \rightarrow \hat{\mathbf{x}}} \frac{1}{|\mathbf{x}|} \sum_{x \in \mathbf{x}} \| x - \phi(x) \|_2,
\end{equation}
where $\phi$ denotes a bijective mapping between $\mathbf{x}$ and $\hat{\mathbf{x}}$.

\section{Channel-adaptive Cross-modal GenSeC-PC}
\label{sec:detailed}
% In this section, we present the design details of the proposed GenSeC-PC framework, which comprises three main parts: cross-modal semantic extraction, channel-adaptive JSCC, and generative semantic inference. 

\subsection{Cross-modal Semantic Extraction}
\begin{figure}
    \centering
    \includegraphics[width=1\linewidth]{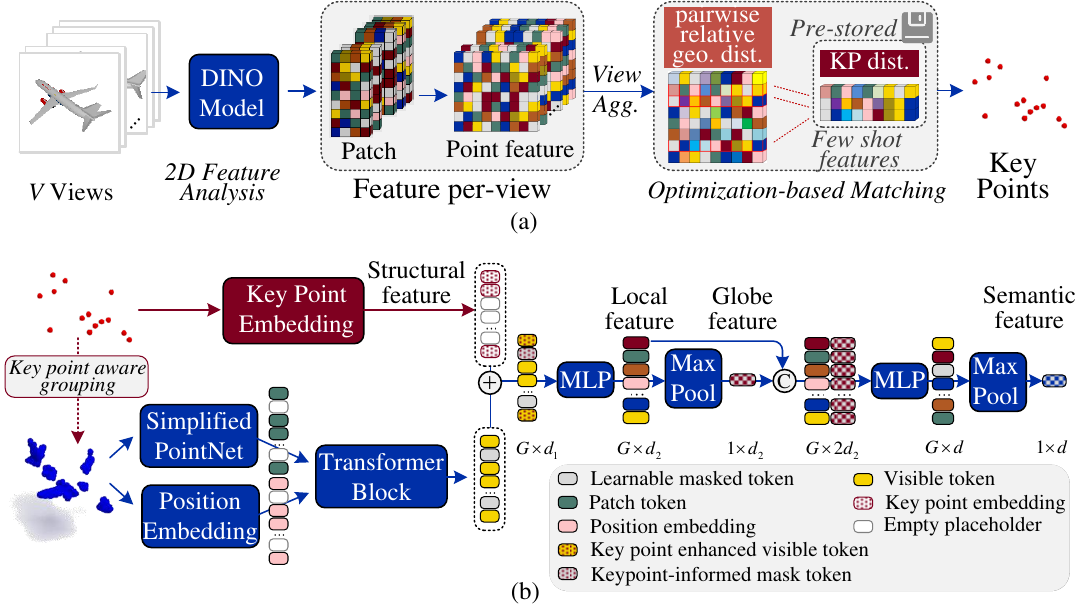}
    \caption{Illustration of cross-modal semantic extraction. (a) Keypoint detection based on B2-3D method form multi-view images; (b) Semantic aggregation and extraction.}
    \label{fig:SE}
    \vspace{-0.5cm}
\end{figure}

Semantic extraction relies on two modalities of data, where PC is dominant, with auxiliary support from image data.  As illustrated in Fig.~\ref{fig:SE}(a), multi-view image data are employed to extract keypoints from the PC, a process that remains highly challenging when attempted solely and directly on the original PC due to the unstructured nature and scarcity of annotated keypoint datasets~\cite{wimmer2023back}. Specifically, images from each view are batched and processed through a pretrained vision encoder (DINOv2~\cite{oquab2024dinov}) to extract visible point features from each projected view. The features are aggregated and back-projected into 3D space to form a comprehensive 3D feature representation. Finally, keypoints in the input PC are identified by matching the few-shot pre-stored keypoint features, associated with the object class, through minimizing both the feature distance and the pairwise relative geodesic distances. 
Nonetheless, a potential limitation of this approach is its reliance on few-shot supervision. It requires the source side to have object detection capabilities and maintain a feature repository for various object categories.
Due to space limitations, more technical details can be found in ~\cite{wimmer2023back}, which are not considered as contributions of this work.

The obtained keypoints outline the structural characteristics of the PC and can be intuitively viewed as the initial extraction of its unique semantic representation. Thus, to more accurately preserve and enhance the robustness of semantic extraction, we fuse the keypoints with the original PC to obtain a richer semantic representation. Specifically, we denote the set of keypoints by $\mathcal{K}$. The total number of keypoints is denoted by $K$, which varies across different objects. The source PC is denoted by ${\bf{x}} \in {\mathbb{R}^{N \times 3}}$ consisting of $N$ points. The $N$ points are first grouped into $G$ groups. To facilitate feature fusion, keypoint-aware farthest point sampling (KP-FPS) is applied, where the keypoints serve as mandatory initial search points. This sampling strategy guarantees that the set of $G$ center points ${\cal C} = \left\{ {{C_i}} \right\}_{i = 1}^G$ includes all keypoints, i.e., ${\cal C} \supseteq \mathcal{K}$. Subsequently, for each center point $C_i$, the $S$ nearest points are retrieved via the K Nearest Neighbors (KNN) algorithm to construct a local point patch $P_i$. To maximize spatial coverage, oversampling is applied such that $G \times S = 2 \times N$~\cite{10901573}. These procedures can be represented as 
\begin{equation}
    \mathcal{C} = {\text{KP-FPS}}\left( {\mathbf{x},\mathcal{K}} \right), \qquad \left\{ {{P_i}} \right\}_{i = 1}^G = \text{KNN}(\mathbf{x},\mathcal{C}).
\end{equation}
To enhance model training stability and accelerate convergence, each point patch undergoes a centering process, where the coordinates of each point are transformed into relative coordinates with respect to the corresponding patch center~\cite{zheng2024point}. Moreover, during the training, we randomly mask parts of the patches with a masking ratio of $\iota$, obtaining visible patches $\left\{ {P_i^{\text{v}}} \right\}_{i = 1}^V$, and masked patches $\left\{ {P_i^{\text{m}}} \right\}_{i = 1}^M$, where $V = \left\lfloor {G \times \iota  } \right\rfloor$ and $M = G - V$. Such a mechanism serves three purposes in GenSeC-PC design: \textit{(1) to encourage the model to develop global reasoning abilities rather than relying on simple memorization, (2) to enhance its robustness in recovering from noisy PCs in real-world applications, and (3) to reduce semantic information content, thereby facilitating more efficient compression coding}. Then, for the visible patches, a simplified PointNet~\cite{qi2017pointnet} ${\xi _\phi }$ with parameter $\phi$ is employed to obtain the patch tokens $\left\{ {T_i^{\text{v}}} \right\}_{i = 1}^V$ with a dimension of $d_1$. Meanwhile, the center point of each visible patch is transformed into position embedding $\left\{ {E_i^{\text{v}}} \right\}_{i = 1}^V$ with the same dimension $d_1$, utilizing a module composed of two linear layers denoted by ${\zeta _\varphi }$. After that, $\left\{ {E_i^{\text{v}}} \right\}_{i = 1}^V$ and $\left\{ {T_i^{\text{v}}} \right\}_{i = 1}^V$ are concatenated and fed into a Transformer Encoder ${\psi _\varpi }$ for extracting latent geometric features and obtaining the final tokens $\left\{ {F_i^{\text{v}}} \right\}_{i = 1}^V$ of visible patches,  which can be expressed by~\cite{10901573} 
\begin{align}
   \left\{ {F_i^{\text{v}}} \right\}_{i = 1}^V = {\psi _\varpi }\left( {{\text{Concat}}\left( {\left\{ {T_i^{\text{v}}} \right\}_{i = 1}^V,\left\{ {E_i^{\text{v}}} \right\}_{i = 1}^V} \right)} \right), \\
    \left\{ {T_i^{\text{v}}} \right\}_{i = 1}^V = {\xi _\phi }\left( {\left\{ {P_i^{\text{v}}} \right\}_{i = 1}^V} \right), \left\{ {E_i^{\text{v}}} \right\}_{i = 1}^V = {\zeta _\varphi }\left( {\left\{ {P_i^{\text{m}}} \right\}_{i = 1}^M} \right).
\end{align}
Moreover,  a set of learnable masked tokens $\left\{ {F_i^{\text{l}}} \right\}_{i = 1}^M$  are initialized for the masked patches. Then, $\left\{ {F_i^{\text{l}}} \right\}_{i = 1}^M$ and $ \left\{ {F_i^{\text{v}}} \right\}_{i = 1}^V$ are interleaved and concatenated into a single sequence following the original order ${\pi _P}$ in $\left\{ {{P_i}} \right\}_{i = 1}^G$. At this point, we obtain the feature of the PC with a dimension $\left( {G \times d} \right)$ as below,
\begin{equation}
\left\{ {F_i^{{\text{c}}}} \right\}_{i = 1}^G = {\text{Concat}}\left( {{\pi _P}\left( {\left\{ {F_i^{\text{v}}} \right\}_{i = 1}^V,\left\{ {F_i^{\text{l}}} \right\}_{i = 1}^M} \right)} \right),{\text{c}} \in \left\{ {{\text{v,l}}} \right\}.
\end{equation}
\begin{figure*}
    \centering
    \includegraphics[width=1\linewidth]{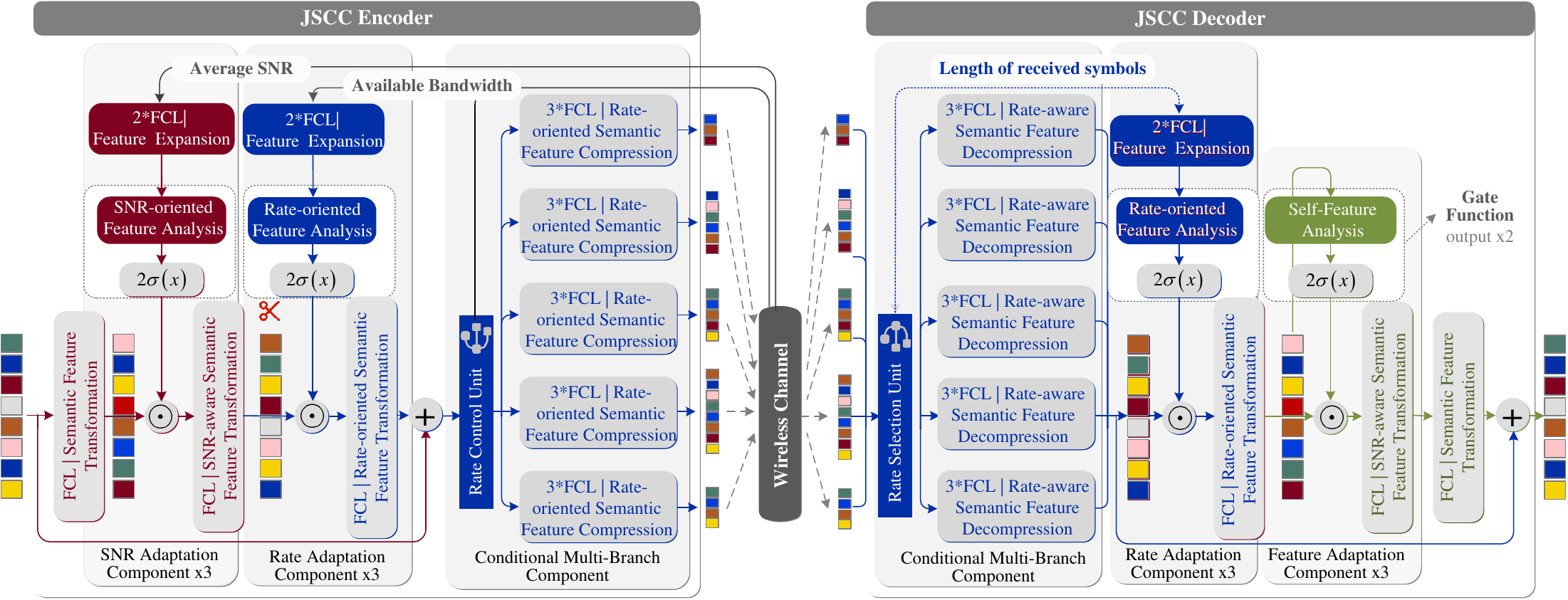}
    \caption{Asymmetric architecture of channel-adaptive JSCC encoder and decoder.}
    \label{fig:JSCC}
    \vspace{-0.3cm}
\end{figure*}

Meanwhile, keypoints are first transformed into keypoint embedding $\left\{ {E_i^{\text{k}}} \right\}_{i = 1}^K$, with the dimension of $d_1$, utilizing a module composed of two linear layers denoted by ${\kappa _\theta }$. Then, to combine the key-point feature and PC feature, the tokens $\left\{ {E_i^{\text{k}}} \right\}_{i = 1}^K$ are added to $\left\{ {F_i^{{\text{c}}}} \right\}_{i = 1}^K$ corresponding to the patches with the keypoints as the center points, irrespective of whether the patch is masked. This can be expressed by 
\begin{equation}
    \tilde F_i^{\text{c}} = \left\{ {\begin{array}{*{20}{c}}
  {F_i^{\text{c}} + E_i^{\text{k}},}&{{\text{if }}i \in \mathcal{K}{\text{ }}} \\ 
  {F_i^{\text{c}},}&{{\text{otherwise}}} 
\end{array}} \right. ,
\end{equation}
where the keypoint embedding plays a role in enhancing the feature representations of the visible token and guiding the learning of mask token. Then, ${\left\{ {\tilde F_i^{\text{c}}} \right\}}_{i = 1}^G$ is fed into a Multi-Layer Perceptron (MLP) module to extract the local features of the PC, where each patch feature is expanded to a dimension of $d_2$. Subsequently, a pooling layer is applied to aggregate the local features with dimensions of $G \times d_2$ into a global feature of dimension $1 \times d_2$. Afterward, the global feature is further expanded, and the concatenation of the global feature and local features is processed by another MLP followed by a pooling layer, resulting in the final semantic feature with a dimension of $1 \times d$, which is denoted by $F_s$ for brevity.

\subsection{Channel-adaptive JSCC}
We propose a novel JSCC scheme, characterized by structural and functional asymmetry. It is derived via pruning from a larger base model with enough capacity to explore the solution, where each adaptive component is conditioned on both the previous feature and the SNR (or rate). The finalized design is shown in Fig.~\ref{fig:JSCC}, and described as below.

The semantic encoder primarily leverages the average SNR and available bandwidth, obtained from channel feedback, to perform two sequential operations: adaptive semantic enhancement tailored to the SNR, followed by rate-oriented semantic compression. 
%This enables the generation of semantic features that are most robust for transmission under the current channel conditions.
Specifically, the semantic feature $F_s$ is first passed through a Fully Connected Layer (FCL) for preliminary feature transformation, obtaining $F_s^0$, followed by $I_{\rm{SNR}}$ repetitions of the SAC. For each SAC $i$, it is composed of a gating function and an FCL. To achieve SNR adaptation, the feedback of the current SNR is first expanded into an SNR feature vector of dimension 
$L$, and then fed into the gating function. 
% The gating function serves to perform fine-grained feature modulation. 
% Meanwhile, as noted in~\cite{10772628}, even under identical channel conditions, different features may experience varying degrees of semantic distortion. Based on this observation, we concatenate the previous feature 
% $F_s^{(i-1)}$ with the SNR feature, and feed the combined vector into the gating function. 
Within the gating function, an FCL perform a further SNR  feature expansion, followed by a sigmoid-based scaling operation whose output is further amplified by a factor of 2 to generate a set of  SNR-adaptive modulation weights. The modulation weights are then applied element-wise to 
$F_s^{(i-1)}$, the input feature of SAC $i$, and the result is passed to the subsequent transformation layer to obtain the updated feature 
$F_s^{(i)}$. The process is repeated $I_1$ times as below,
 \begin{equation}
     W_{\left| {1 \times d} \right|}^i = 2 \times {\rm{Gate}}_{\rm{E}}^{\rm{S}}\left( {{\rm{FC}}{{\left( {{\rm{SNR}}} \right)}_{\left| {1 \times L} \right|}}} \right),
 \end{equation}
 \begin{equation}
     {F_s^{(i)}}_{\left|1 \times d \right|} = {\text{FC}}_{\rm E}^{\rm S}\left( {{W_{\left|1 \times d \right|}^i} \odot {F{{_s^{\left( {i - 1} \right)}}_{\left|1 \times d \right|}}}} \right).
 \end{equation}
Afterwards, the SNR-adaptive feature   $F_s^{(I_1)}$ is fed into the rate-oriented semantic compression module, which operates in two stages. The first stage comprises ${I_{\rm{rate}}^{\rm E}}$ RACs.  RAC is designed to facilitate pre-aggregation of features for rate-adaptive compression.  This design aims to strike a good trade-off between model redundancy and accuracy. The specific process is similar to SAC, which can be outlined as
\begin{equation}
    W_{\left|1 \times d \right|}^{{I_{\rm{SNR}}} + i} = {\text{2}} \times {\text{Gate}_{\rm E}^{\rm R}}\left( {{{\text{FC}}{{\left( {{\text{rate}}} \right)}_{\left|1 \times L \right|}}} } \right),
\end{equation}
\begin{equation}
    {F_s^{({I_{\rm{SNR}}} + i)}}_{\left|1 \times d \right|} = {\text{FC}}_{\rm E}^{\rm R}\left( {W_{\left|1 \times d \right|}^{{I_{\rm{SNR}}} + i} \odot F{{_s^{\left( {{I_{\rm{SNR}}} + i - 1} \right)}}_{\left|1 \times d \right|}}} \right).
\end{equation}
Finally, leveraging residual learning, we incorporate a skip connection from the input to the feature $F_s^{({I_{\rm{SNR}}}+{I_{\rm{rate}}^{\rm E}})}$ to   accelerate convergence and simplify training objectives. Then, the resultant sum is then processed by a CMBC, which can be expressed by 
\begin{equation}
    {S_{\text{T}}}_{\left| {1 \times r} \right|} = {\text{CMBC}}_{\rm E}\left[ r \right]\left\{ {{\text{F}}{{\text{C}}^{(r)}}\left( {F{{_s^{({I_{\rm{SNR}}} + {I_{\rm{rate}}^{\rm E}})}}_{\left|1 \times d \right|}}} \right) + {F_s}} \right\},
\end{equation}
where $r \in \mathcal{R}$ denotes the target rate and $\mathcal{R}$ denotes the set of available rates, and $S_{\text{T}}$ denotes the transmission symbol.

\begin{figure}
    \centering
\includegraphics[width=0.9\linewidth]{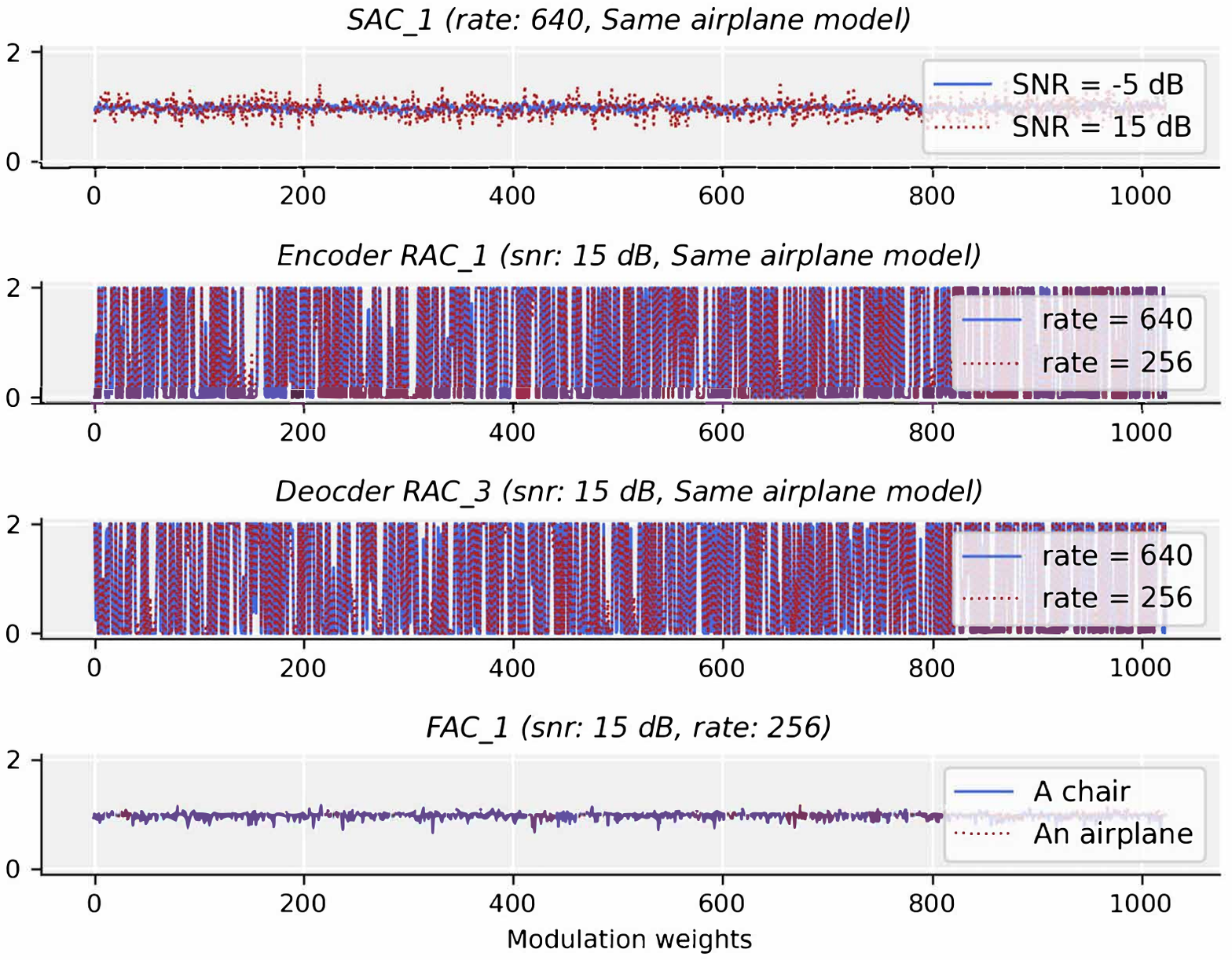}
    \caption{Modulation weights of each adaption component.}
    \label{fig:modulation}
    \vspace{-0.6cm}
\end{figure}

After compression, the semantic features are modulated via amplitude modulation to generate analog signals, which are subsequently transmitted over the wireless channel. We consider the general fading channel model with transfer
function ${{\hat S}_{\text{R}}} = {\mathbf{h}} \odot {S_{\text{T}}} + {\mathbf{n}}$, where ${{\hat S}_{\text{R}}}$ denotes the received feature, $\mathbf{h}$ denotes the channel gain and ${\mathbf{n}}$ denotes the additive noise vector. The elements of ${\mathbf{n}}$ are independently drawn from a Gaussian distribution, i.e., ${\mathbf{n}} \sim \mathcal{N}\left( {0,\sigma _n^2{\mathbf{I}}} \right)$,  where $\sigma _n^2$
is the average noise power. Without loss of generality, we consider both the additive white Gaussian noise (AWGN) channel and the Rayleigh fading channel in this work.

Different from the semantic encoder, once the receiver obtains the semantic feature $\hat{S}_{\text{R}}$, the JSCC decoder performs two different operations.
The first is rate-adaptive semantic decompression, which mirrors the encoding process in reverse,  utilizing a CMBC and $I^{\rm D}_{\rm{rate}}$ RACs. Moreover, as noted in~\cite{10772628}, even under identical channel conditions, different features may experience varying degrees of semantic distortion. To address this, the second operation aims to achieve self-feature adaptation with $I_{\rm{Fea}}$ FACs, aiming to mitigate the differential impact of transmission-induced semantic loss across diverse semantic features.
Specifically, after determining the length of $\hat{S}_{\text{R}}$, the CMBC is employed to decompress $\hat{S}_{\text{R}}$ based on its rate $r$, thereby recovering the feature $\hat{F}_s^{{I_{\rm{SNR}}}+{I_{\rm{rate}}^{\rm E}}}$, which corresponds to $F_s^{{I_{\rm{SNR}}}+{I_{\rm{rate}}^{\rm E}}}$ at the sender side and can be expressed by
\begin{equation}
    {\hat{F}{{_s^{({{I_{\rm{SNR}}}+{I_{\rm{rate}}^{\rm E}}})}}_{\left|1 \times d \right|}}}={\text{CMBC}}_{\rm D}\left[ r \right]\left\{ {{\text{F}}{{\text{C}}^{(r)}}\left( {\hat{S}{_{\text{R}}}}_{\left| {1 \times r} \right|}  \right)} \right\},r \in \mathcal{R}.
\end{equation}
Then, ${{\hat F}{{_s^{({{I_{\rm{SNR}}}+{I_{\rm{rate}}^{\rm E}}})}}_{\left|1 \times d \right|}}}$ is  fed into ${I^{\rm D}_{\rm{rate}}}$ RACs sequentially, which is similar to the RACs at the encoder, outlined as 
\begin{equation}
    W_{\left|1 \times d \right|}^{{I_{\rm{Fea}}} + i} = {\text{2}} \times {\text{Gate}_{\rm D}^{\rm R}}\left( {{{\text{FC}}{{\left( {{\text{rate}}} \right)}_{\left|1 \times d \right|}}} } \right),
\end{equation}
\begin{equation}
    \hat{F}{{_s^{\left( {{I_{\rm{Fea}}} + i } \right)}}_{\left|1 \times d \right|}} = {\text{FC}}_{\rm D}^{\rm R}\left( {W_{\left|1 \times d \right|}^{{I_{\rm{Fea}}} + i} \odot \hat{F}{{_s^{\left( {{I_{\rm{Fea}}} + i + 1} \right)}}_{\left|1 \times d \right|}}} \right).
\end{equation}

Then, the obtained $\hat{F}{_s^{\left( {{I_{\rm{Fea}}}} \right)}}$ is fed into   $I_{\rm{Fea}}$ FACs in order. FAC is similar to SAC and RAC, except that the SNR or rate feature is replaced by the input feature $F_s^{(i+1)}$ of FAC $i$. It can be highlighted as
\begin{equation}
    W_{\left|1 \times d \right|}^{ i} = {\text{2}} \times {\text{Gate}_{\rm D}^{\rm S}}\left( { {\hat{F}{{_s^{\left( { i + 1} \right)}}_{\left|1 \times d \right|}}} } \right),
\end{equation}
\begin{equation}
    \hat{F}{{_s^{\left( { i } \right)}}_{\left|1 \times d \right|}} = {\text{FC}}_{\rm D}^{\rm S}\left( {W_{\left|1 \times d \right|}^{ i} \odot \hat{F}{{_s^{\left( { i + 1} \right)}}_{\left|1 \times d \right|}}} \right).
\end{equation}
At last, the  feature $\hat{F}_s^0$ is fed into the final FCL to reconstruct the  feature with a skip connection ${\hat F}{{_s^{({{I_{\rm{SNR}}}+{I_{\rm{rate}}^{\rm E}}})}}}$ as 
\begin{equation}
  \hat{F}_s =   {\text {FC}}\left(\hat{F}_s^0\right) + {\hat F}{{_s^{({{I_{\rm{SNR}}}+{I_{\rm{rate}}^{\rm E}}})}}}.
\end{equation}
The impact of the main components involved in JSCC is visualized in Fig.~\ref{fig:modulation}.

\subsection{Diffusion-based Semantic Inference}
\begin{figure}
    \centering
\includegraphics[width=1\linewidth]{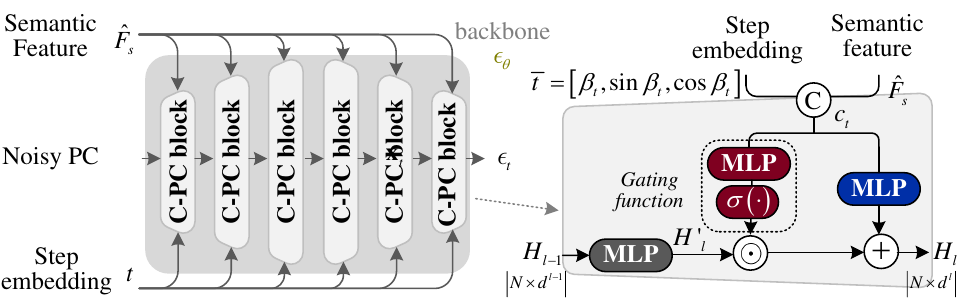}
    \caption{Backbone of the PC diffusion.}
    \label{fig:diffutionbackbone}
    \vspace{-0.2cm}
\end{figure}
The semantic decoder is built upon a controlled PC Diffusion model as its core. In contrast to general diffusion process with the particle positions becoming increasingly randomized, which can be modeled as a Markov chain
\begin{equation}
    q\left( {{{\mathbf{x}}^{1:T}}\left| {{{\mathbf{x}}^0}} \right.} \right) = \coprod\limits_{t = 1}^T {q\left( {{{\mathbf{x}}^t}\left| {{{\mathbf{x}}^{t - 1}}} \right.} \right)}, \label{eq:FP}
\end{equation}
where  ${q\left( {{{\mathbf{x}}^t}\left| {{{\mathbf{x}}^{t - 1}}} \right.} \right)}$  is Markov diffusion kernel~\cite{luo2021diffusion}, the decoding process can be interpreted as the reverse of this diffusion. Specifically, $\mathbf{x}^0$ is  equivalent to $\mathbf{x}$ at sources, and $\mathbf{x}^T$ is the pure noisy PC. Moreover,
$q\left( {{{\mathbf{x}}^t}\left| {{{\mathbf{x}}^{t - 1}}} \right.} \right) = \mathcal{N}\left( {{{\mathbf{x}}^t};\sqrt {1 - {\beta _t}} {{\mathbf{x}}^{t - 1}},{\beta _t}{\mathbf{I}}} \right)$, where the hyperparameters ${{\beta _t}}$ are  some pre-defined small constants and gradually increase. Moreover, according to~\cite{ho2020denoising}, ${{{\mathbf{x}}^t}}$ can be elegantly formulated as a direct function of ${{{\mathbf{x}}^0}}$,
\begin{equation}
    q\left( {{{\mathbf{x}}^t}\left| {{{\mathbf{x}}^0}} \right.} \right) = \mathcal{N}\left( {{{\mathbf{x}}^t};\sqrt {{{\bar \alpha }_t}} {{\mathbf{x}}^0},\left( {1 - {{\bar \alpha }_t}} \right){\mathbf{I}}} \right), \label{eq:ss}
\end{equation}
where ${{\bar \alpha }_t} = \prod\nolimits_{i = 1}^t {{\alpha _i}}$ and ${\alpha _t} = 1 - {\beta _t}$. Similar to \eqref{eq:FP}, the reverse process can also be modeled as a Markov chain. The difference lies in that
the reverse process aims to recover the desired shape from
the input noise, which requires training from sampled data relying on ~\eqref{eq:ss}. Specifically, the reverse process controlled by $\hat{F}_s$ can be expressed by 
\begin{equation}
    {p_\theta }\left( {{{\mathbf{x}}^{0:T}}\left| {{{\hat F}_s}} \right.} \right) = p\left( {{{\mathbf{x}}^T}} \right)\prod\limits_{t = 1}^T {{p_\theta }\left( {{{\mathbf{x}}^{t - 1}}\left| {{{\mathbf{x}}^t},\hat F_s} \right.} \right)}, 
\end{equation}
where ${p_\theta }\left( {{{\mathbf{x}}^{t - 1}}\left| {{{\mathbf{x}}^t},\hat F_s} \right.} \right) = \mathcal{N}\left( {{{\mathbf{x}}^{t - 1}};{\mu _\theta }\left( {{{\mathbf{x}}^t},t,\hat F_s} \right),\sigma _t^2{\mathbf{I}}} \right)$, and ${\mu _\theta }$ is the trainable mean relying on the backbone as shown in Fig.~\ref{fig:diffutionbackbone}. Specifically, the backbone for PC diffusion, denoted by ${\epsilon _\theta }$, consists of multiple controlled-PC (C-PC) blocks~\cite{luo2021diffusion}, following a Decoder-Encoder architecture that structurally resembles an ``inverse U-Net”.  For each C-PC block, the feature from previous block $H_{l-1}$ is first fed into an MLP, performing either dimensional expansion or compression, obtaining $H'_l$. Meanwhile, the step $t$ is transformed into an embedding $\bar t = \left[ {{\beta _t},\sin {\beta _t},\cos {\beta _t}} \right]$. Then, it is concatenated with semantic feature $\hat{F}_s$, obtaining the combined condition for each step $c_t$.   The resulting $c_t$
  is simultaneously fed into a gating function to obtain the modulation weight ${W_t^l}$ for $H'_l$, and into an MLP to generate the bias term. The above can be outlined as 
\begin{equation}
    {c_t} = {\text{Concat}}\left( {{{\hat F}_s},\bar t} \right),
\end{equation}
\begin{equation}
    {H_l} = {\text{Gate}}\left( {{c_t}} \right) \odot {\text{MLP}}\left( {{H_{l - 1}}} \right) + {\text{MLP}}\left( {{c_t}} \right).
\end{equation}
The training method for the PC diffusion is detailed in Section~\ref{sec:train_methodology}.  Moreover, inspired  by DDIM~\cite{song2020denoising},  to accelerate the inference process, a subsequence $\left[ {{\tau _1}, \ldots ,{\tau _S}} \right]$ of the original diffusion steps $\left[ {1, \ldots ,T} \right]$ can be selected, resulting in a shorter forward diffusion trajectory. Accordingly, the reverse  process can be expressed by
\begin{equation}
\begin{aligned}
    {{\mathbf{x}}^{{\tau _{i - 1}}}} = \sqrt {{{\bar \alpha }_{{\tau _{i - 1}}}}} \left( {\frac{{{{\mathbf{x}}^{{\tau _i}}} - \sqrt {1 - {{\bar \alpha }_{{\tau _{i - 1}}}}} {\epsilon_\theta }\left( {{{\mathbf{x}}^{{\tau _i}}},{\hat F_s},{\tau _i}} \right)}}{{\sqrt {{{\bar \alpha }_{{\tau _i}}}} }}} \right) \\+ \sqrt {1 - {{\bar \alpha }_{{\tau _{i - 1}}}} - \sigma _{{\tau _i}}^2} {\epsilon_\theta }\left( {{{\mathbf{x}}^{{\tau _i}}},{\hat F_s
    },{\tau _i}} \right) + {\sigma _{{\tau _i}}},
    \end{aligned}
\end{equation}
where ${\sigma _{{\tau _i}}}$ can be set to 0, if deterministic sampling is adopted.
\begin{figure}
    \centering
\includegraphics[width=1\linewidth]{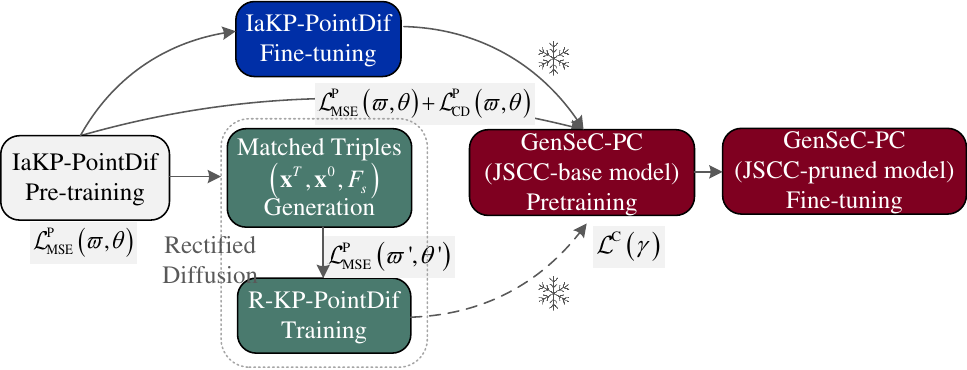}
    \caption{Training flow of GenSeC-PC.}
    \label{fig:trainingflow}
    \vspace{-0.6cm}
\end{figure}
\subsection{Training Methodology}
\label{sec:train_methodology}
The training flow of the overall GenSeC-PC is shown in Fig.~\ref{fig:trainingflow}. Firstly, we perform the pre-training for the  PC diffusion with cross-modal semantic extraction (called image-assisted keypoint based (IaKP)-PointDif for short), which only includes the semantic encoder and semantic decoder. It is trained  following the same  methodology from~\cite{zheng2024point}, with the exception of using a smaller batch size. This choice is motivated by our experimental observations that a reduced batch size leads to improved generalization and higher accuracy of the model. We denote the whole semantic encoder by ${\mathcal{E}_\omega  }\left(  \cdot  \right)$, with the whole trainable parameter $\omega$. Following the simplified version of the MSE derived in~\cite{ho2020denoising}, the loss function during the pretraining can be expressed by
\begin{equation}
\mathcal{L}_{{\text{MSE}}}^{\text{P}}\left( {\omega ,\theta } \right) = {\text{MSE}}\left( {\epsilon, {\epsilon_\theta }\left( {\sqrt {\bar \alpha } {{\mathbf{x}}^0} + \sqrt {1 - \bar \alpha } \epsilon,{\mathcal{E}_\omega }\left( {{{\mathbf{x}}^0}} \right),t} \right)}  \right).
\end{equation}
Then, given the unstructured and unordered nature of PC, we add the CD metric during the fine-tuning process, which can be represented as 
\begin{equation}
\mathcal{L}_{{\text{CD}}}^{\text{P}}\left( {\omega ,\theta } \right) = {\text{CD}}\left( {\epsilon, {\epsilon_\theta }\left( {\sqrt {\bar \alpha } {{\mathbf{x}}^0} + \sqrt {1 - \bar \alpha } \epsilon,{\mathcal{E}_\omega }\left( {{{\mathbf{x}}^0}} \right),t} \right)}  \right).
\end{equation}
Thus, the final loss function for fine-tuning can be represented by $\mathcal{L}_{{\text{MSE}}}^{\text{P}}\left( {\omega ,\theta } \right) + \mathcal{L}_{{\text{CD}}}^{\text{P}}\left( {\omega ,\theta } \right)$. Subsequently, we freeze the parameters of IaKP-PointDif and insert the JSCC encoder, wireless channel, and decoder into the pipeline. During this training phase, we jointly optimize the JSCC encoder and decoder following pretraining-pruning pipeline by considering two metrics: MSE between the input feature $F_s$ to the JSCC encoder and the reconstructed feature $\hat{F}_s$
  from the JSCC decoder, and the  CD between the input ${\mathbf{{x}}}$
 to the semantic encoder and the final output ${\mathbf{\hat{x}}}$
  from the semantic decoder. Thus, the loss function for the training of the JSCC parameter  can be expressed by 
\begin{equation}
    {\mathcal{L}^{\text{C}}}\left( \gamma  \right) = {\text{MSE}}\left( {{F_s},{{\hat F}_s}} \right) + {\text{CD}}\left( {{\mathbf{x}},{\mathbf{\hat x}}} \right),
\end{equation}
Up to now, a well-trained GenSeC-PC framework has been constructed.
To accelerate the training process and reduce memory consumption, DDIM sampling is employed to derive ${\mathbf{\hat{x}}}$ during the training.
However, to mitigate the potential performance degradation introduced by the accelerated DDIM sampling, we incorporate RD~\cite{wang2025rectified} into the training phase.
The core idea of RD is to obtain a first-order approximate ordinary differential equation path, thereby enabling the model to follow a smoother and more data-consistent trajectory.
The key to achieving this lies in generating matched noise-sample pairs using a pre-trained diffusion model and retraining the model on these paired data.
Specifically, the pre-trained IaKP-PointDif is used to generate triplets consisting of the initial noise PC ${{\mathbf{x}}^T}$, the target PC ${{\mathbf{x}}^0}$, and the condition $F_s$.
Based on the generated dataset, we then re-train the IaKP-PointDif model, which we denote as R-IaKP-PointDif.
Subsequently, the original IaKP-PointDif module in the GenSeC-PC framework can be replaced with the rectified R-IaKP-PointDif,   achieving superior performance with   few  sampling steps compared to directly applying DDIM on the original IaKP-PointDif.
\section{Simulation and Evaluation}
\label{sec:simulation}
\subsection{Simulation Setup}
We use KeypointNet~\cite{9157559} for pretraining and fine-tuning the GenSeC-PC model, and evaluate it on both KeypointNet and ShapeNet~\cite{chang2015shapenet}. It contains 8,329 3D models across 16 categories, each with a varying number of keypoints. Following~\cite{zheng2024point}, we sample 1024 points based on FPS algorithm  for each 3D model to serve as the source data, i.e., $N = 1024$.  Moreover, the number of group is set to  $G = 64$, the size of each group is set to $S= 32$, and the mask ratio is set to $\iota = 0.8$. The feature dimensions during the semantic extraction process are set to $d_1 = 384$, $d_2 = 512$, $d = 1024$. In the design of the JSCC codec, the SAC, the RACs at the encoder and decoder and the FAC are configured with three stages. The backbone of PC diffusion consists of 6 C-PC blocks. The output dimensions $H_l$ for C-PC block are [128, 256, 512, 256, 128, 3].  The total denoising step is set to $T = 2000$, and $\beta_t$ linearly increases from 1e-4 to 1e-2. The length of the embedding for SNR and rate is set to $L=128$.
\begin{figure*}
    \centering
    \includegraphics[width=1\linewidth]{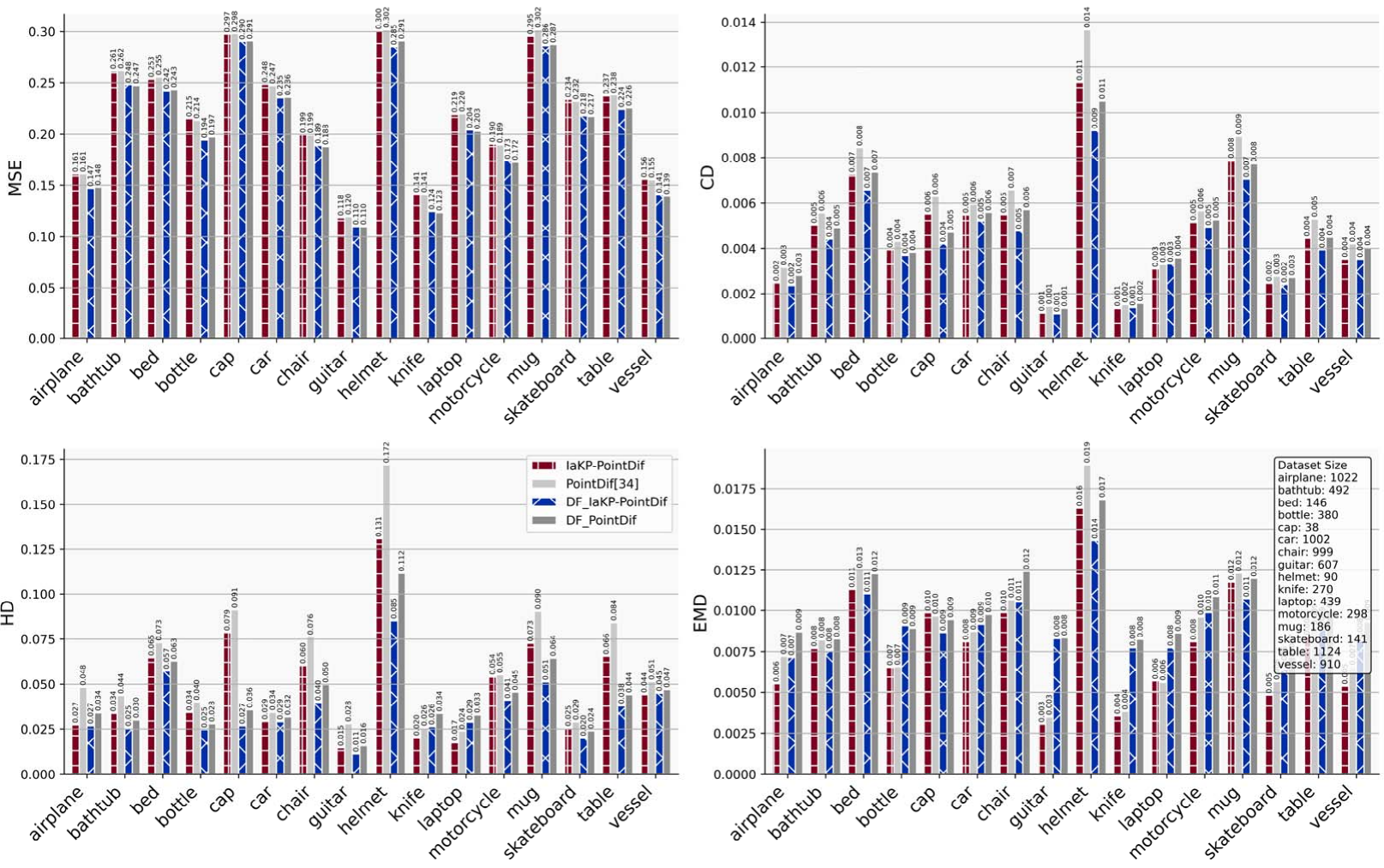}
    \caption{Performance comparison between IaKP-PointDif and PointDif.}
    \label{fig:Kp-compraison}
\end{figure*}
\begin{figure*}
    \centering
    \includegraphics[width=0.9\linewidth]{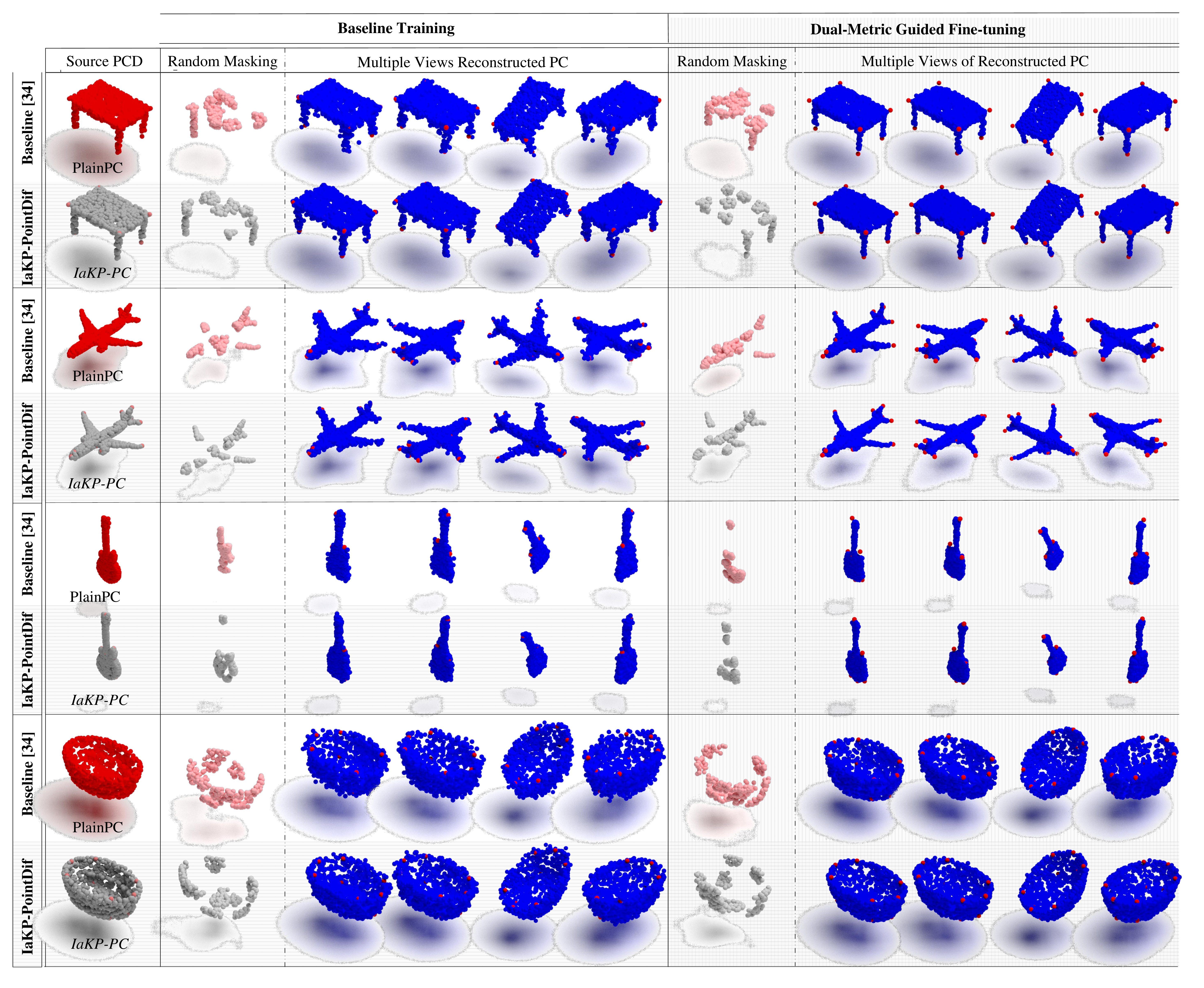}
    \caption{Comparison of visualization results between IaKP-PointDif and PointDif.}
    \label{fig:V-Kp-compraison}
\end{figure*}
To enhance generalization during the training, data augmentation is performed using random scaling and translation. Moreover, during the pre-training, IaKP-PointDif is trained for 500 epochs.  We adopt the AdamW optimizer with a weight decay of 0.05 and a learning rate of 0.0006. We apply the cosine decay schedule to adjust the learning rate with 10 warm-up epochs. For fine-tuning, IaKP-PointDif is trained for 100 epochs using AdamW with a weight decay of 0.0001 and a learning rate of 0.0001. Moreover, the batch size for both pre-trainig and fine-tuning is set to 32. To improve learning efficiency, we divide the interval 
$\left( {0,T} \right]$ into 12 equal segments. During each training iteration, for a given input, the model randomly samples a time step from each segment. During the training for JSCC codec, we define the possible rate levels as [1024, 896, 768, 640, 512, 384, 256, 128, 64, 32] and SNR levels as [-15 dB, -10 dB, -5 dB, 0 dB, 5 dB, 10 dB, 15 dB, 20 dB]. We adopt the AdamW optimizer with a weight decay of 0.0001 and a learning rate of 0.001. The JSCC codec is first trained on the base model for 300 epochs, followed by pruning and an additional 300 epochs of fine-tuning. Each stage includes 10 warm-up epochs. The batch size is set to 16. In each batch, the SNR is randomly sampled and a common   rate is applied to all samples. Across batches, both the SNR and the rate are independently and randomly determined. Furthermore, we utilize the pre-trained IaKP-PointDif model to generate 560,000 sampled triples for RD training, following the same training methodology as used in IaKP-PointDif pre-training. \textit{The code and well-trained model can be established at \url{https://github.com/wty2011jl/GenSeC-PC}}.

\subsection{Performance Analysis} 
The performance analysis is conducted in three stages. 
% \textit{First}, ablation experiments  confirm that our semantic extraction and dual-metric fine-tuning improve performance over PointDif~\cite{zheng2024point}. \textit{Second}, we perform end-to-end communication simulations based on the GenSeC-PC framework and compare the results with those of traditional communication schemes, demonstrating the robustness and transmission efficiency of our method. Meanwhile, we evaluate the effectiveness of DDIM and RD strategies in reducing decoding latency. \textit{Lastly}, we verify the robustness of GenSeC-PC against keypoint noise and unseen  shapes.

% we evaluate the proposed framework under two challenging scenarios: (1) based on existing B2-3D~\cite{wimmer2023back} algorithms, we assess its performance when keypoint detection is inaccurate; and (2) using the ShapeNet dataset, we verify the robustness of the framework when handling previously unseen models during training, including both typical and uncommon shapes.
\subsubsection{Ablation on Semantic Extraction}

To evaluate the impact of the proposed cross-modal semantic extraction and  fine-tuning strategy with the dual-metric $(\mathcal{L}_{{\text{MSE}}}^{\text{P}}\left( {\omega ,\theta } \right) + \mathcal{L}_{{\text{CD}}}^{\text{P}}\left( {\omega ,\theta } \right))$, we conduct experiments without considering wireless channel effects. We compare the original PointDif~\cite{zheng2024point}, the proposed IaKP-PointDif, as well as the fine-tuned versions of both models. Notably, each pair is trained using the same methodology to ensure a fair comparison. Figure.~\ref{fig:Kp-compraison} illustrates the reconstruction performance of all object categories in the dataset using the four aforementioned models, evaluated across four different metrics. 
First, from an overall perspective, all four models are capable of generating objects across multiple categories, although the reconstruction accuracy varies among categories. This variation may be attributed to differences in object complexity, object size, and the number of samples per category in the dataset. Moreover, it is evident that for both the pre-trained and fine-tuned models, the IaKP-PointDif-based approach consistently achieves better reconstruction accuracy across all four evaluation metrics compared to the PointDif-based models. The improvements are particularly pronounced in the more point-cloud-specific similarity metrics—CD, HD, and EMD. A similar trend can be observed from a different perspective: for both the PointDif-based and IaKP-PointDif-based approaches, fine-tuning leads to significant improvements in reconstruction performance, especially with respect to CD and HD across most object categories.   In this context, both the cross-modal semantic extraction and the dual-metric-based fine-tuning are shown to have their respective positive contributions. 
Moreover, visualization results of the four models are presented in Fig.~\ref{fig:V-Kp-compraison}. To better illustrate the reconstruction accuracy of the 3D objects, we show four different viewpoint projections for each generated objects. Moreover, rows labeled `P' represent the pre-trained model, while rows labeled `F' correspond to the fine-tuned model.
The visual results indicate that IaKP-PointDif achieves more accurate reconstruction of fine structural details than PointDif. Notable improvements include the definition of table corners, the consistency of table leg thickness, and the precise representation of airplane vertices. Moreover, the IaKP-PointDif outputs contain significantly fewer outlier points, contributing to cleaner overall reconstructions. Moreover, the fine-tuned models generally produce objects with better visual quality compared to their pre-trained counterparts, exhibiting clearer surfaces and contours as well as fewer outlier points. \textit{However, fine-tuning method may also introduces slight changes in object size, and in some cases, leads to overcompensation effects.} Thus, as shown in Fig.~\ref{fig:Kp-compraison}, although the fine-tuned models achieve better performance in CD and HD metrics, they may exhibit a slight decline in reconstruction accuracy under the EMD metric for certain object categories. Therefore, in scenarios where point-to-point precision is not strictly required and visual similarity suffices, the fine-tuned models are more suitable. Conversely, for applications demanding high geometric accuracy, the pre-trained models may be preferable.
\subsubsection{Performance  Evaluation of GenSeC-PC}
\begin{figure*}[htbp]
  \centering

  \begin{subfigure}[b]{0.23\linewidth}
    \includegraphics[width=\linewidth]{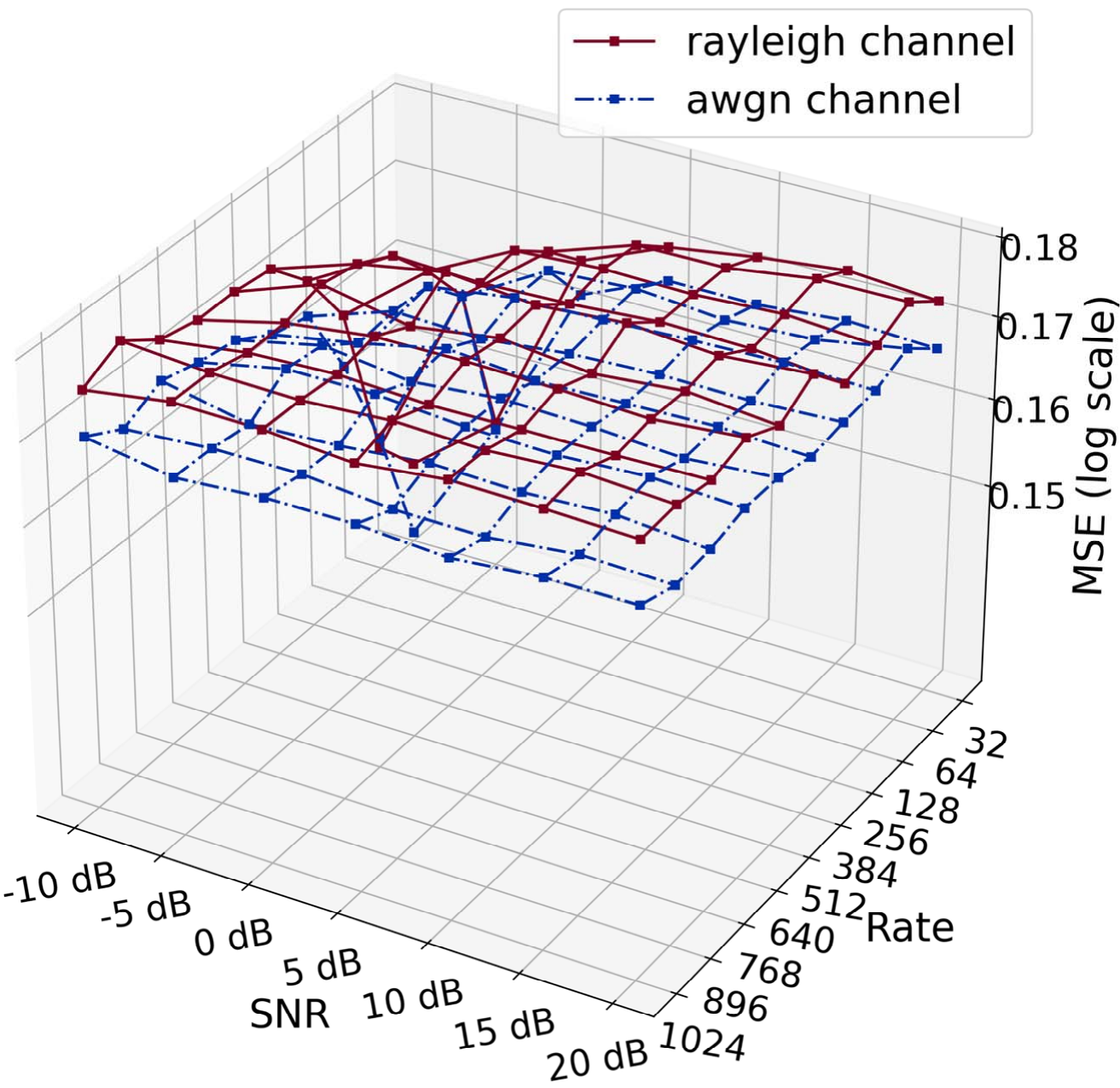}
  \end{subfigure}
  \hfill
  \begin{subfigure}[b]{0.23\linewidth}
    \includegraphics[width=\linewidth]{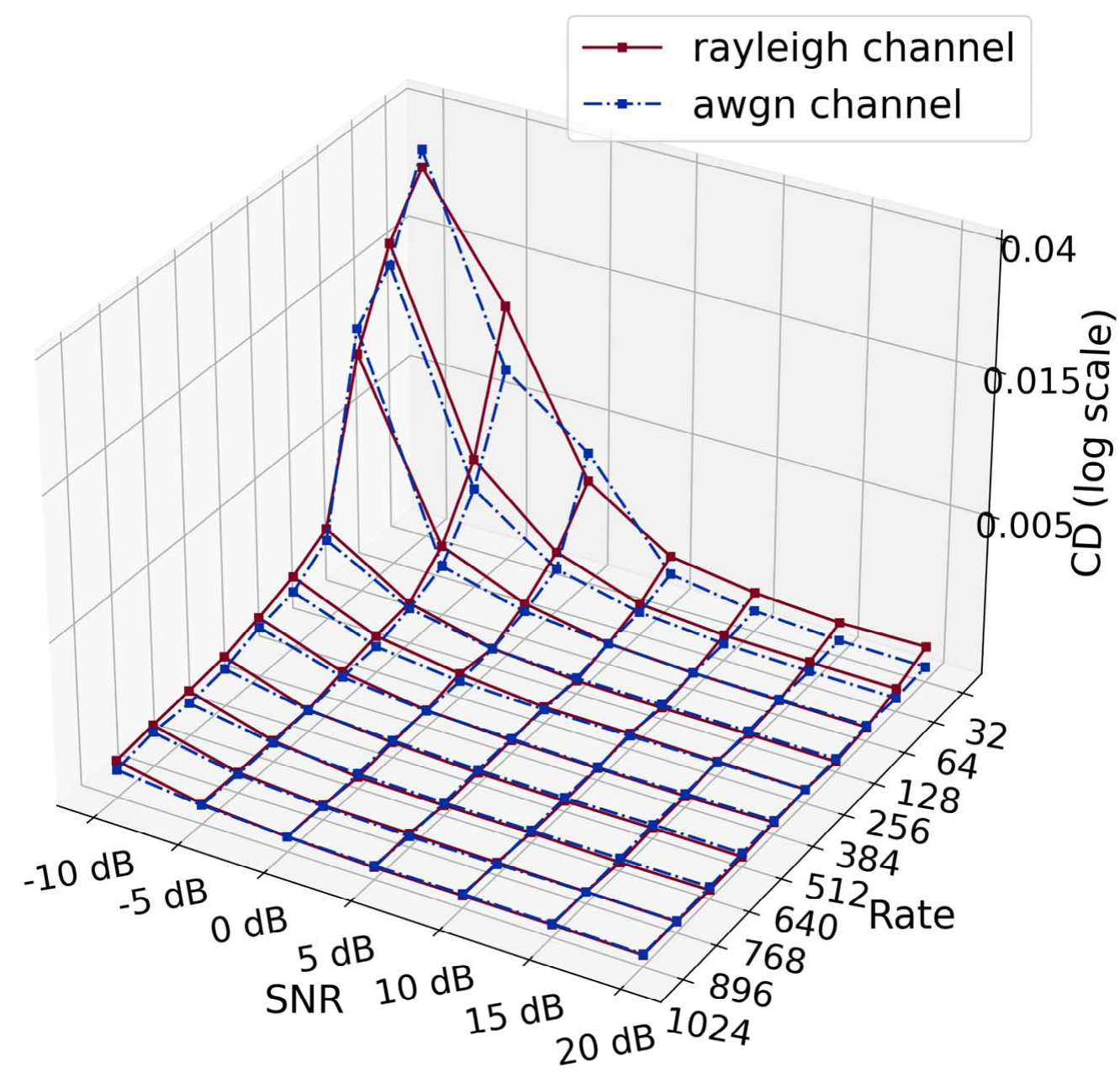} 
  \end{subfigure}
  \hfill
  \begin{subfigure}[b]{0.23\linewidth}
    \includegraphics[width=\linewidth]{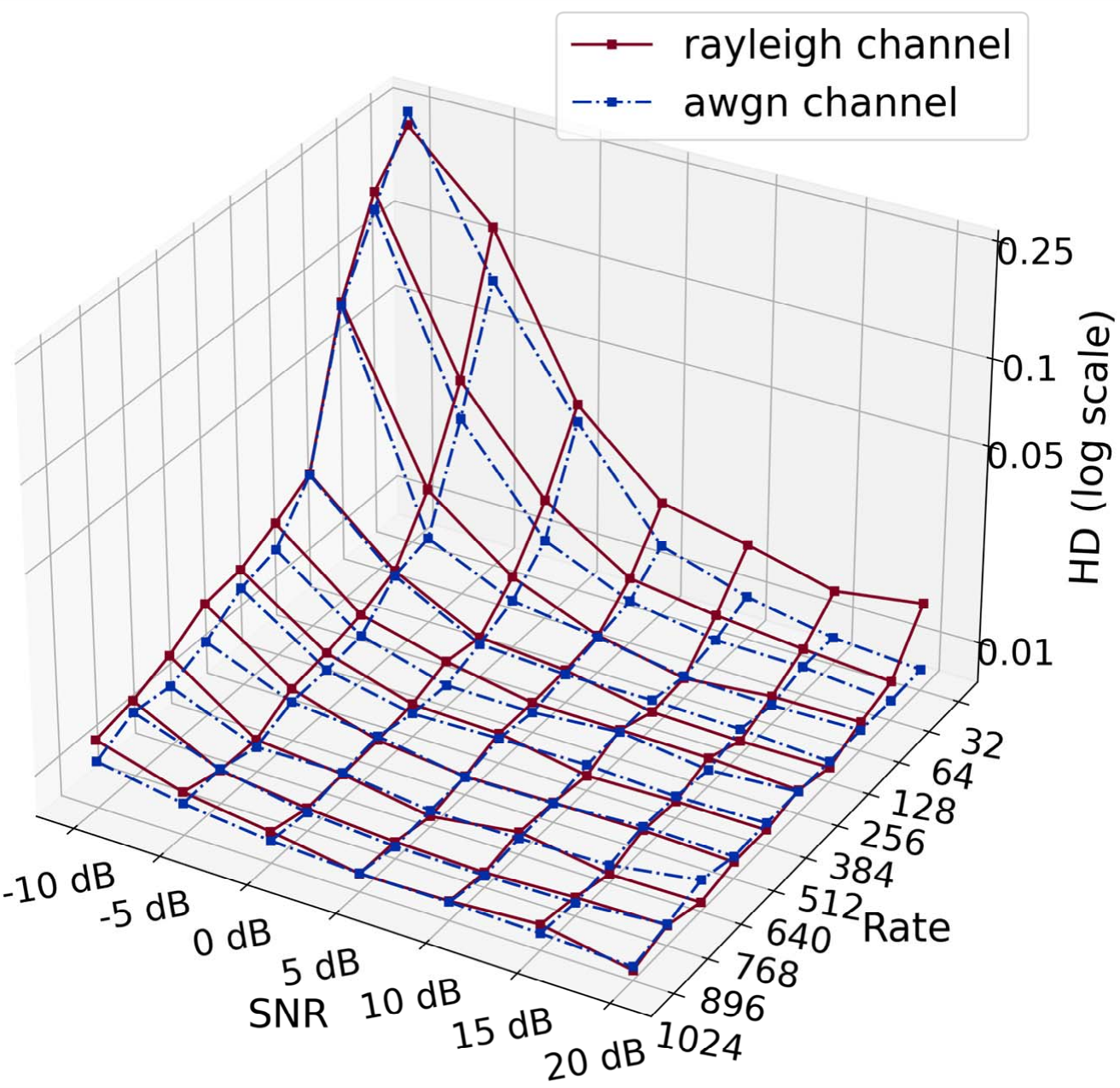}
  \end{subfigure}
  \hfill
  \begin{subfigure}[b]{0.23\linewidth}
    \includegraphics[width=\linewidth]{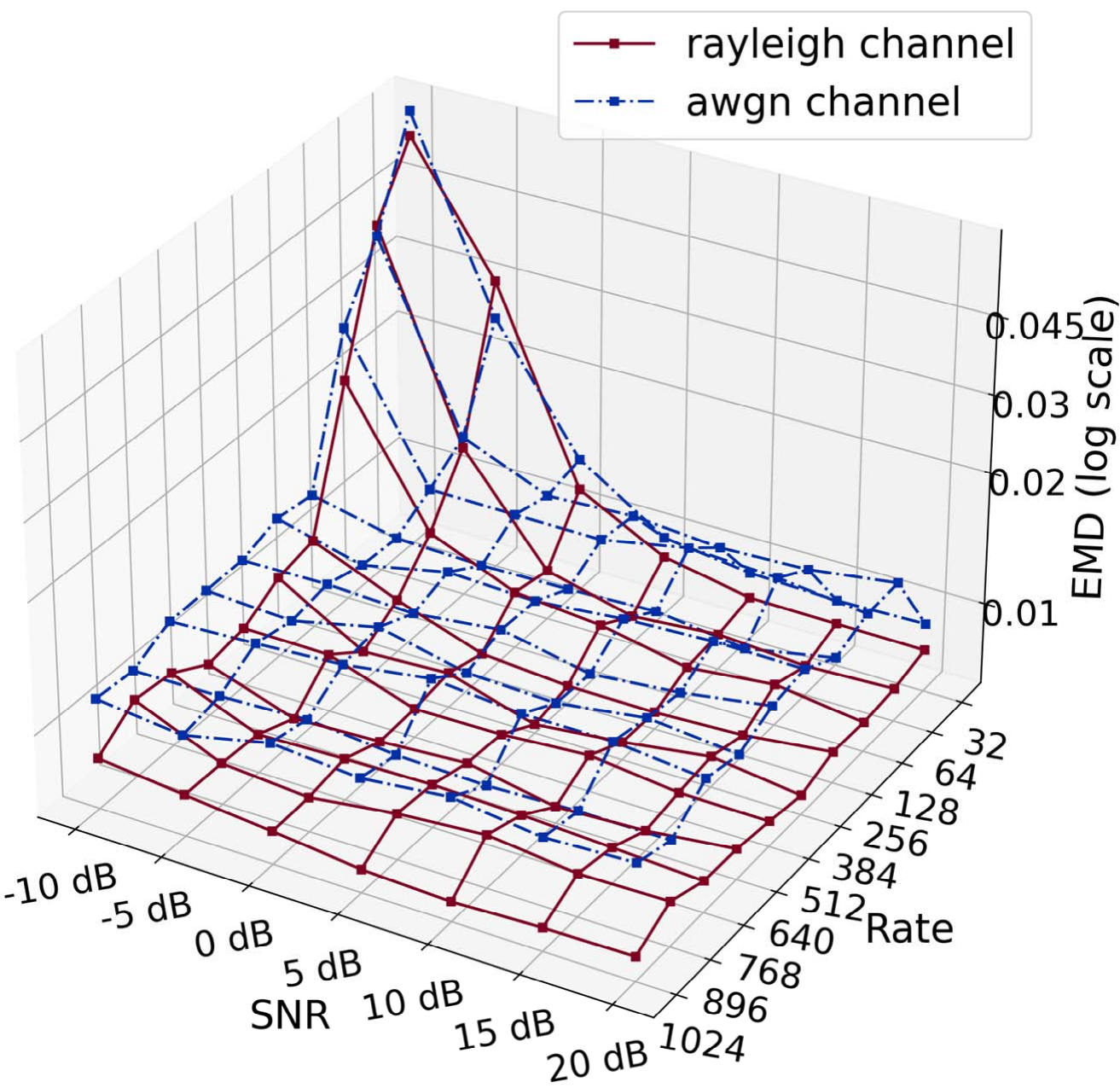}
  \end{subfigure}

  \caption{Communication performance of GenSeC-PC under AWGN and Rayleigh channels with varying SNR and rate.}
  \label{fig:AWGN}
\end{figure*}

In the following experiments, we adopt the fine-tuned IaKP-PointDif model as the semantic encoder-decoder for the GenSeC-PC framework. To isolate the impact of object-level variability on the evaluation metrics, we select a single object and transmit it 100 times under different combinations of SNR and rate. The mean and variance of each metric are then computed  and shown in Figs.~\ref{fig:AWGN}, where the colored surface represents the mean values, and two semi-transparent gray surfaces are used to indicate the variance range. As shown in the two figures, except for the MSE metric, which fails to capture PC similarity due to the unordered and unstructured nature of PC, the other three metrics, CD, HD, and EMD, consistently decrease as SNR and rate increase. Moreover, as expected, the performance under the AWGN channel is generally better than that under the Rayleigh fading channel. An interesting observation is that across various SNR conditions, the communication performance at a rate of 1024 is consistently inferior to that at 896. This indicates that effectively aggregating and compressing features before re-expansion may contribute to improved robustness, as opposed to directly transmitting higher-dimensional representations. 
\begin{figure}
    \centering
\includegraphics[width=1\linewidth]{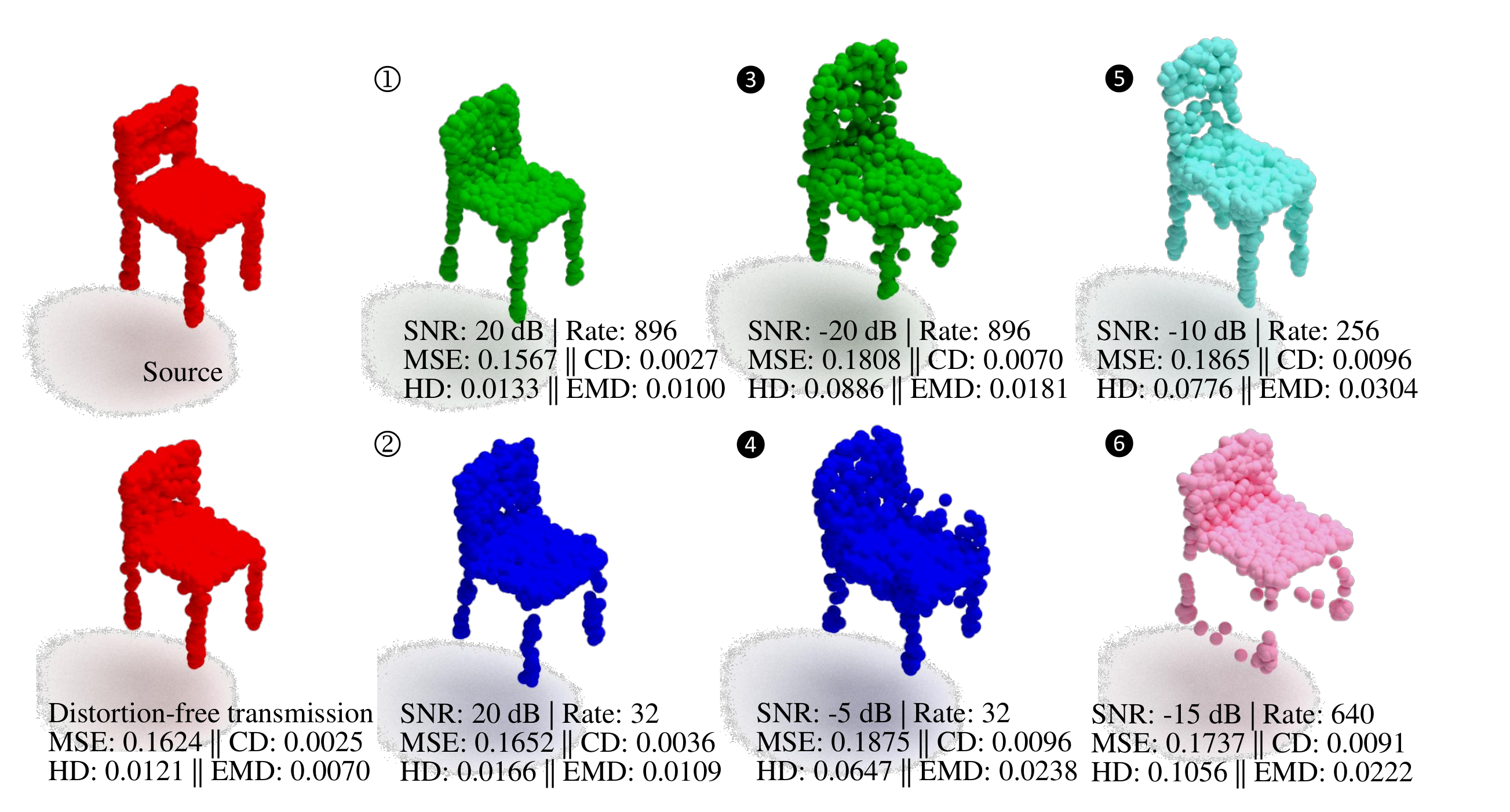}
    \caption{Typical visualization of transmission results under Rayleigh channel.}
    \label{fig: visualexample1}
\end{figure}
\begin{figure}
    \centering
\includegraphics[width=1\linewidth]{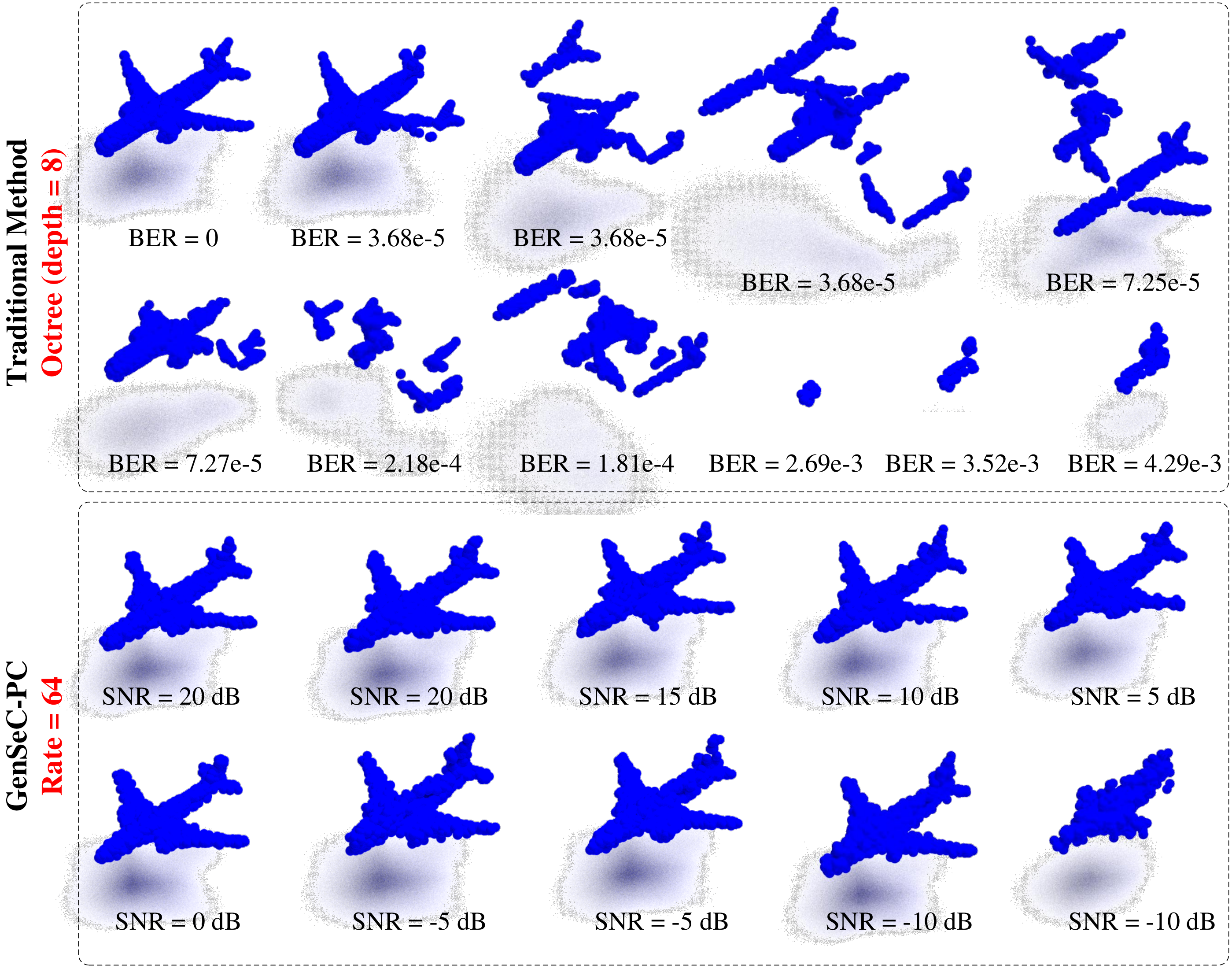}
    \caption{Visual comparison between GenSeC-PC and conventional communication under varying transmission conditions.}
    \label{fig: visualexample2}
\end{figure}
Some typical visual  transmission results are shown in Fig.~\ref{fig: visualexample1}. As illustrated in the figure, under favorable channel conditions (e.g., SNR = 20 dB), both rate = 896 and rate = 32 achieve performance comparable to distortion-free transmission. However, each rate level corresponds to a different minimum acceptable channel quality. In general, higher rates yield stronger transmission robustness. Specifically, at rate = 896, the model is able to reconstruct the general structure of the source PC even at SNR = -20 dB, albeit with noticeable distortions. In contrast, for rate = 32, a similar degradation in reconstruction quality is observed at SNR = -5 dB. Likewise, the critical SNR points for rates of 640 and 265 are approximately -15 dB and -10 dB, respectively. Moreover, under these extremely low SNR conditions, various types of distortions may occur in the reconstructed PCs. In some cases, the points become overly dispersed, leading to blurred object contours—for example, the ``chair" in Examples 3 and 4 loses its well-defined edges. In other instances, although the overall shape of the object changes significantly, as seen in Examples 5 and 6, the reconstructed PC still resemble chairs. This behavior differs significantly from that of traditional communication methods, which is  detailed  below.
\begin{figure*}[t]
  \centering

  \begin{subfigure}[b]{0.32\linewidth}
    \includegraphics[width=\linewidth]{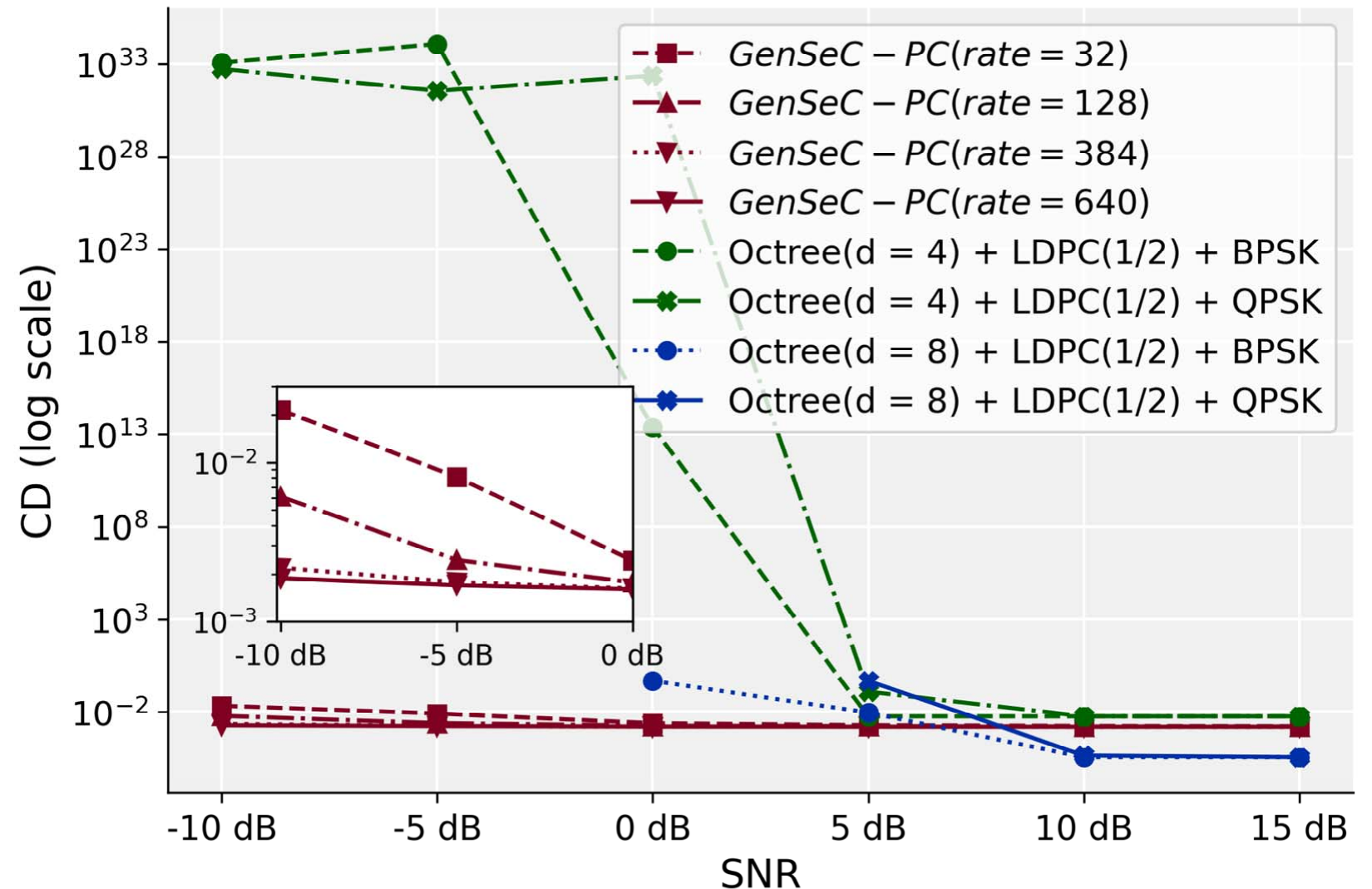}
    \caption{}
  \end{subfigure}
  \begin{subfigure}[b]{0.32\linewidth}
    \includegraphics[width=\linewidth]{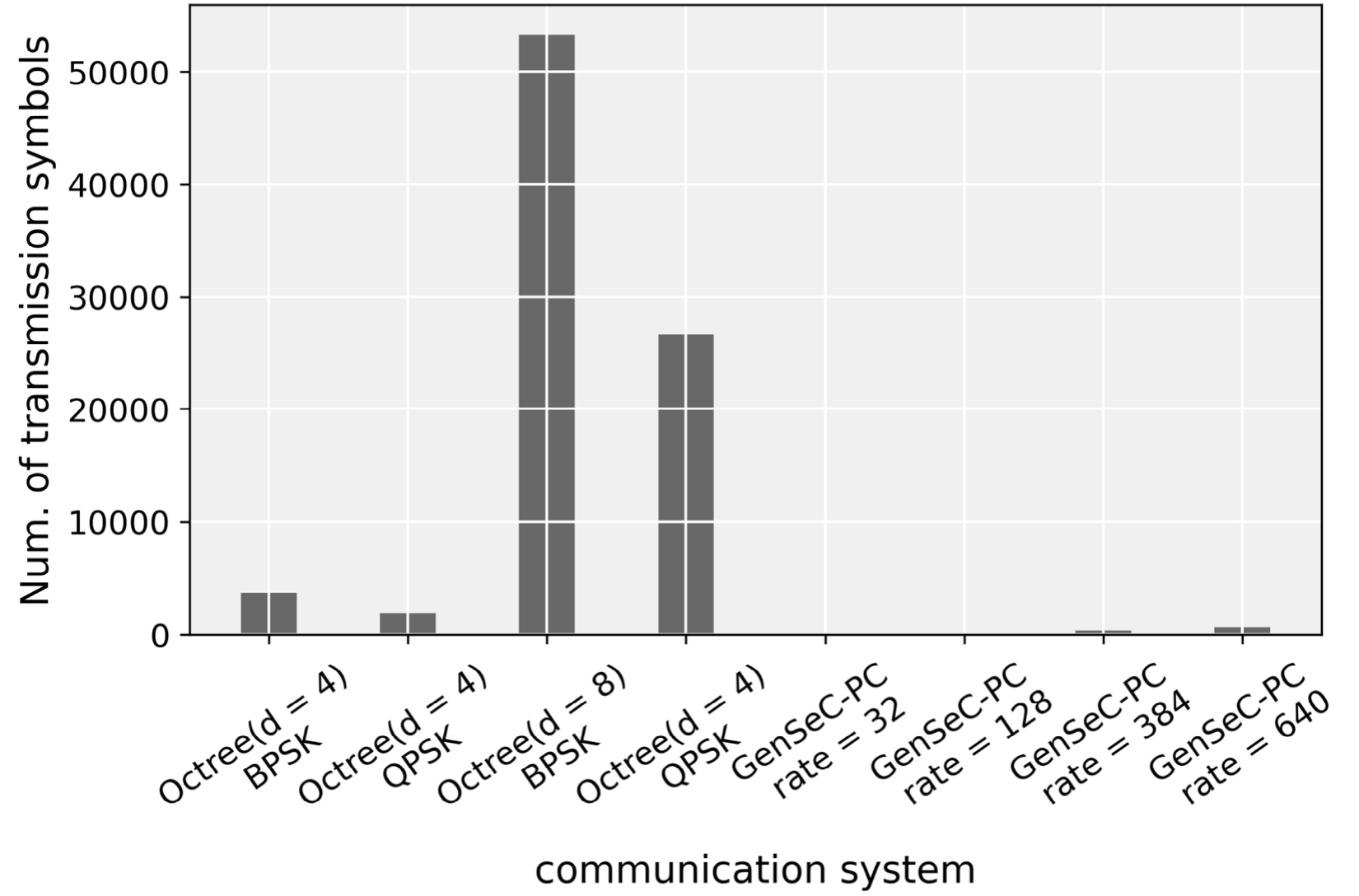}
    \caption{}
  \end{subfigure}
  \begin{subfigure}[b]{0.32\linewidth}
    \includegraphics[width=\linewidth]{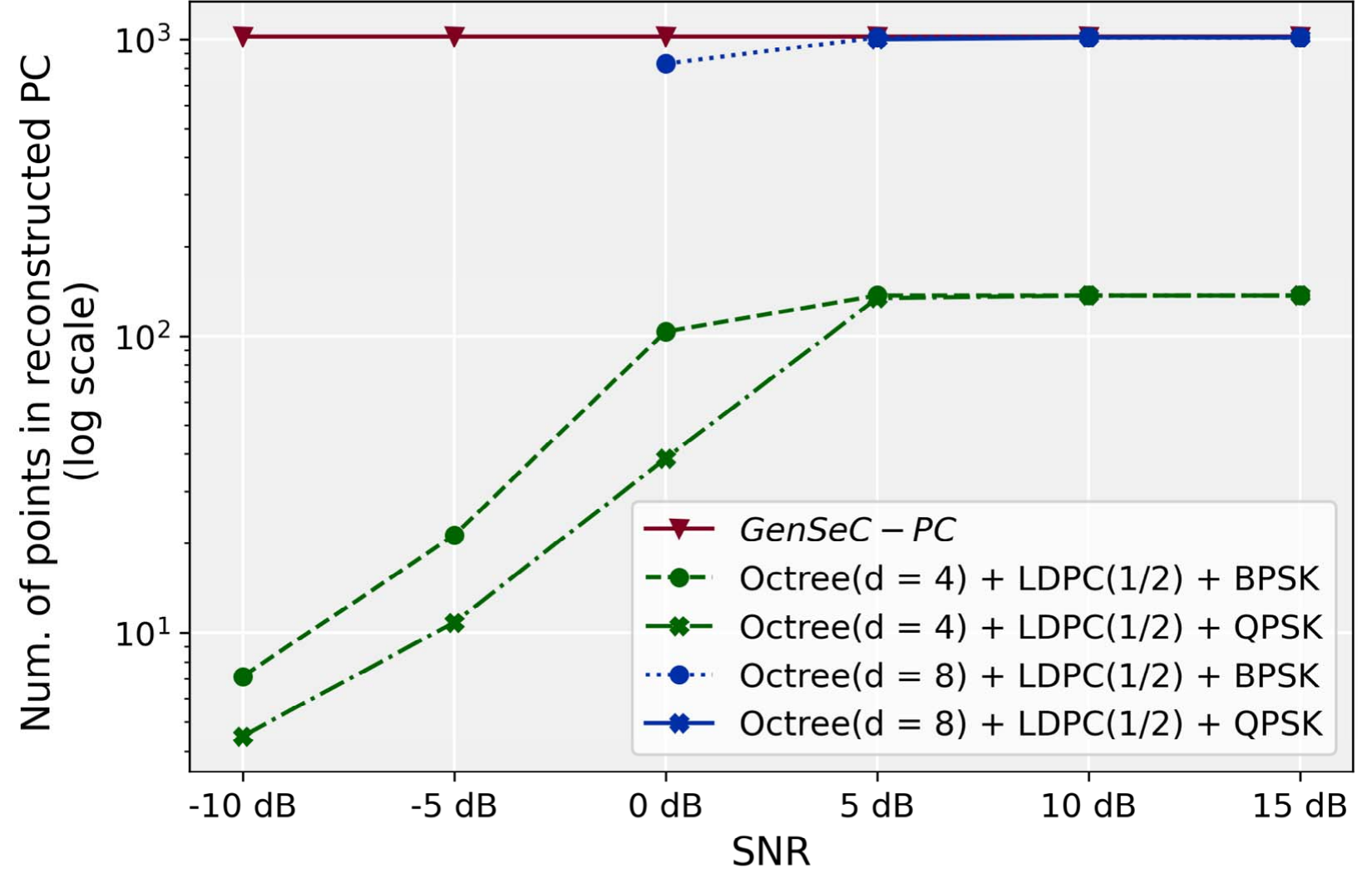}
    \caption{}
  \end{subfigure}

  \caption{Performance comparison between GenSeC-PC and traditional method under AWGN channel, averaged over 100 independent transmissions.}
  \label{fig:CC}
\end{figure*}
\begin{figure}[t]
    \centering
\includegraphics[width=1\linewidth]{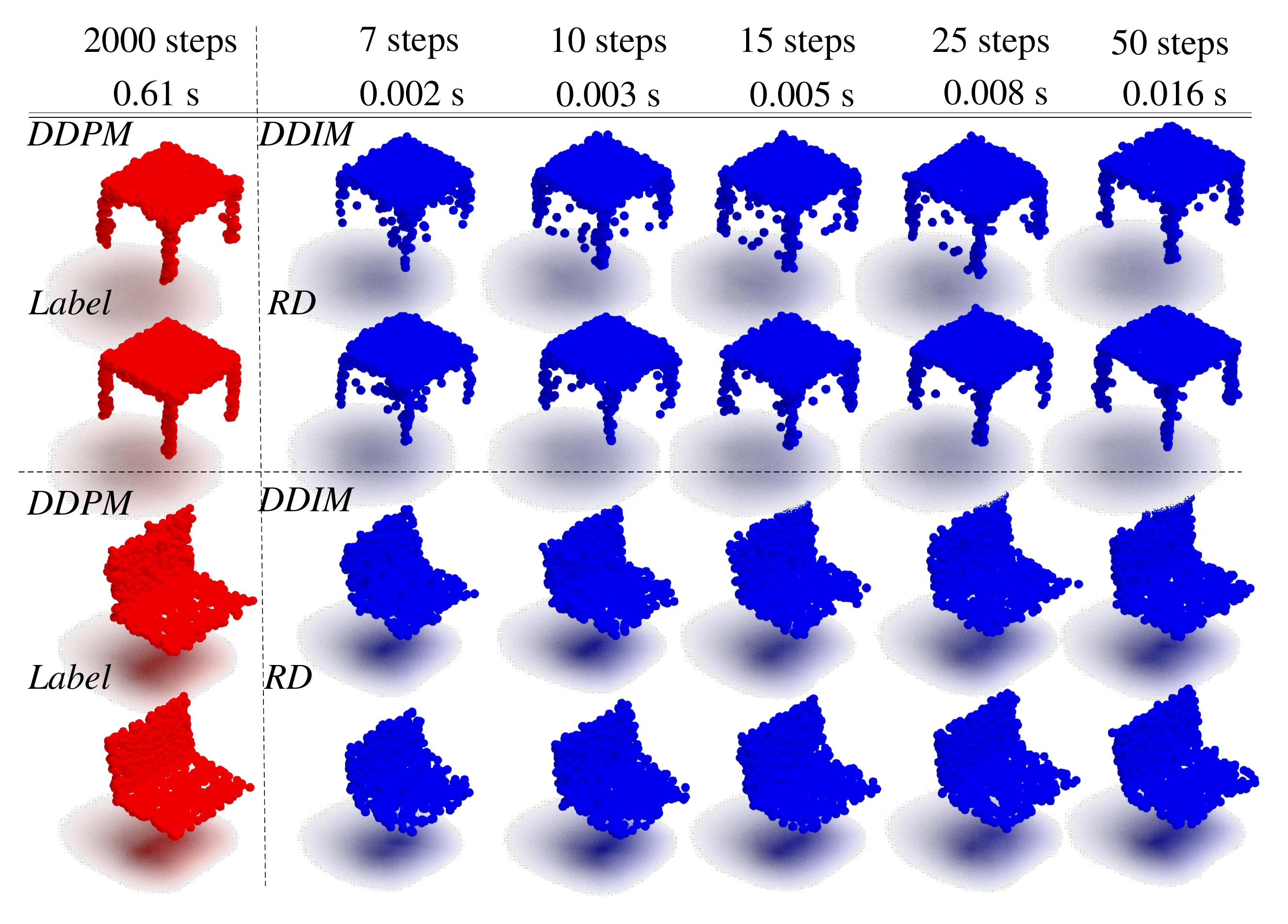}
    \caption{Decoding acceleration with DDIM and RD.}
    \label{fig: decoding}
\end{figure}
\begin{figure}[t]
    \centering
\includegraphics[width=1\linewidth]{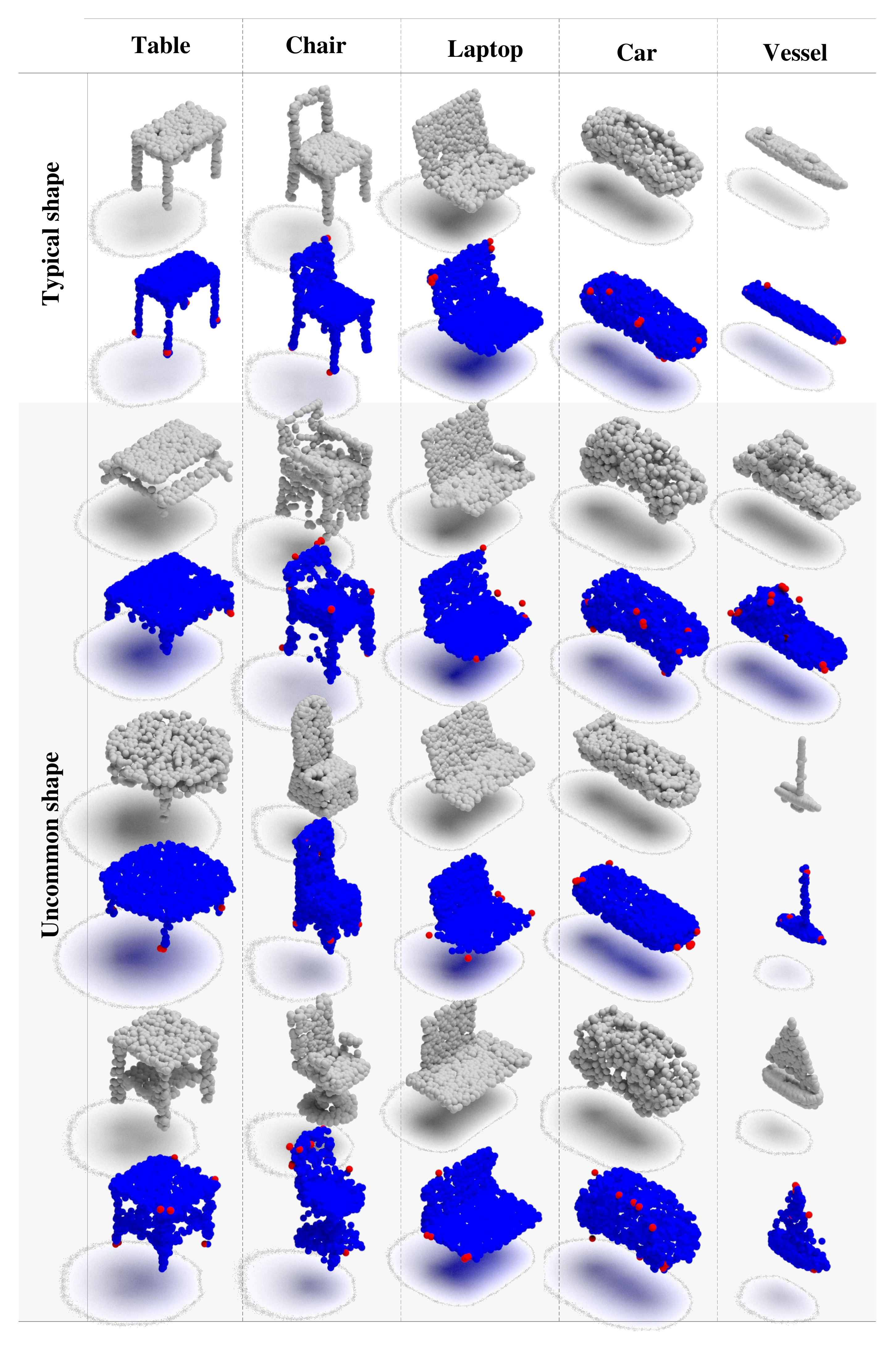}
    \caption{Performance evaluation on the unseen models in ShapeNet dataset with SNR = 15 dB and rate = 512.}
    \label{fig: shapenet_V}
\end{figure}

Figure~\ref{fig: visualexample2} compares the transmission distortion of GenSeC-PC and a conventional communication method based on Octree encoding (depth = 8) under different channel conditions. As shown in the figure, even a single bit error in the conventional method can lead to unpredictable and severe deformation of the PC shape. In contrast, GenSeC-PC consistently preserves the overall shape across various channel conditions. This demonstrates the remarkable robustness of GenSeC-PC against transmission-induced errors.
Subsequently, we conduct a comprehensive comparison between GenSeC-PC and conventional methods under various settings, in terms of point cloud reconstruction quality, bandwidth usage, and the number of generated points. To ensure reliable transmission, we employ a rate-1/2 LDPC channel code together with BPSK or QPSK modulation. As shown in Fig.~\ref{fig:CC}(a), the conventional method exhibits a clear ``cliff effect".  When the Octree encoding depth is set to 8, decoding fails entirely under low-SNR conditions. Although reducing the depth to 4 enables successful decoding, the resulting distortion remains extremely high. In contrast, the proposed GenSeC-PC achieves significantly lower CD across various SNR and rate settings.
Furthermore, Fig.~\ref{fig:CC}(b) illustrates the number of transmitted symbols, where GenSeC-PC requires substantially less bandwidth—over 1000× fewer symbols in some cases. It is also worth noting that although reducing the Octree depth reduces the transmitted code size, it comes at the cost of significantly degraded reconstruction accuracy. Even under lossless transmission, Octree decoding still results in higher CD compared to GenSeC-PC, likely due to insufficient point density caused by the complexity of the object shape as conveyed by Fig.~\ref{fig:CC}(c).

\begin{figure*}
    \centering
    \includegraphics[width=1\linewidth]{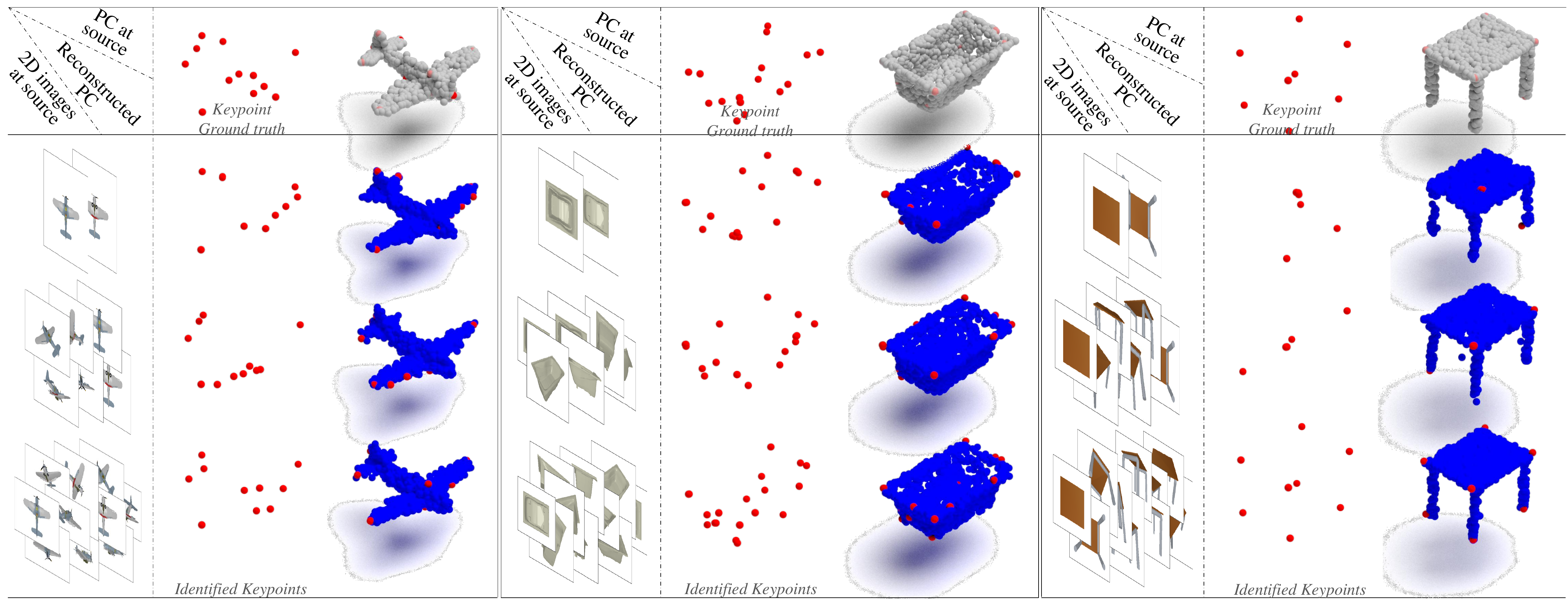}
    \caption{Visualization results of keypoint detection, using different numbers of view images for acquisition and PC reconstruction at a rate of 512  per cloud under an SNR of 15 dB.}
    \label{fig: Vrobustness}
\end{figure*}
\begin{figure}[t]
    \centering
\includegraphics[width=1\linewidth]{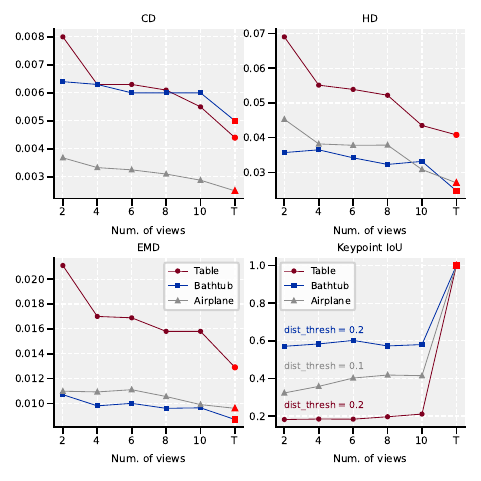}
    \caption{Average performance of keypoint detection, using varying numbers of view images for acquisition, and PC reconstruction accuracy with   rate = 512  and SNR = 15 dB.}
    \label{fig: robust_curve}
\end{figure}
Next, we validate the effectiveness of using DDIM and RD for accelerating inference in PC diffusion. The computational latency in this section is tested using an RTX 4090 GPU. As shown in Fig.~\ref{fig: decoding}, both DDIM and RD can significantly reduce the inference steps and decoding latency of the original DDPM, making it suitable for most real-time communication scenarios in practical applications, despite a slight decrease in reconstruction accuracy. Moreover, compared to DDIM, using RD results in fewer scattered points and a more defined object edge, even when reducing to the same number of steps. This confirms the necessity of re-training the decoding process based on RD in scenarios with high latency requirements. However, it is important to note that training RD requires generating triples based on a pre-trained model, which then uses a rectify flow to improve generation speed. Since we modified the original data distribution during the dual-metric fine-tuning process, the application of RD is limited to scenarios involving the pretrained IaKP-PointDif model.

\subsubsection{Robustness Evaluation of GenSeC-PC}

We evaluate the proposed framework under two challenging scenarios:
1) based on existing B2-3D~\cite{wimmer2023back} algorithms, we assess its performance under inaccurate keypoint detection;
2) using the ShapeNet~\cite{chang2015shapenet} dataset, we evaluate its robustness to previously unseen models, including both typical and uncommon shapes.
Fig.~\ref{fig: Vrobustness} presents keypoint detection and PC reconstruction results at a transmission rate of 512 points per cloud and an SNR of 15 dB, using varying numbers of input view images. Viewpoint positions are uniformly distributed across the sphere using Fibonacci sphere sampling. As the number of 2D views increases, keypoint detection becomes more accurate, leading to improved reconstruction detail. Despite occasional missing or misidentified keypoints, the framework consistently preserves the overall structure of the reconstructed model, demonstrating strong robustness to keypoint detection errors.
Fig.\ref{fig: robust_curve} shows the average reconstruction performance and keypoint detection accuracy (measured by IoU\cite{wimmer2023back}) across the first 100 shapes in three categories. ``T" indicates ground-truth keypoints. The trends in the curves support the observations from the visual results.
Furthermore, using the acquisition of 10-view 2D images as an example, we simulate the performance of GenSeC-PC on unseen shapes from the ShapeNet dataset under an SNR of 15 dB and a rate of 512. We select one typical shape and several uncommon ones from representative object categories—table, chair, laptop, car, and vessel—to evaluate robustness. As shown in Fig.~\ref{fig: shapenet_V}, the model achieves reconstruction quality comparable to training samples for typical shapes, even with unseen variations (e.g., height/width differences in tables and chairs, size and opening angles of laptops, or structural differences in cars and vessels). For complex unseen shapes with rich details, the framework still reconstructs the overall structure accurately, with minor deviations in fine details.

\section{Conclusion}
\label{sec:conclusion}
In this paper, we have proposed a novel channel-adaptive cross-modal GenSeC-PC framework for  PC reconstruction. Specifically, we have designed a cross-modal semantic encoder that fuses features from both 2D images and 3D PCs, enabling a more comprehensive semantic representation. Additionally, we have introduced a pair of streamlined and asymmetric channel-adaptive JSCC encoders and decoders, conditioned on both SNR and available bandwidth, which effectively balance redundancy and accuracy—without requiring any additional overhead. To address the decoding latency inherent in traditional diffusion models, we have incorporated RD and DDIM into the inference process, thereby meeting the real-time demands of communication systems. Finally, we have conducted end-to-end training and transmission evaluations under various channel models and conditions using the KeypointNet and ShapeNet datasets. An ablation study has further validated the effectiveness of the proposed cross-modal semantic extraction module and dual-metric guided fine-tuning. Simulation results confirm the effectiveness of cross-modal semantic extraction, highlighting the framework’s robustness across diverse conditions—including low SNR, bandwidth limitations, varying numbers of input images, and previously unseen objects based on ShapeNet dataset.

While our method shows strong performance, several limitations remain, mainly due to the B2-3D approach. First, DINOv2 is effective but too heavy for embedded deployment, highlighting the need for lighter alternatives. Second, although the storage cost of the feature dictionary is acceptable in most scenarios, it implies that the sensor must be equipped with object detection capabilities. Finally, since reconstruction quality depends on keypoint accuracy, improving detection performance remains an important future direction.

\bibliographystyle{IEEEtran}
\bibliography{ref}

% \cite{}
% \section*{Biographies}
% \small

% {Author} is with

\end{document}